\documentclass[lettersize,journal]{IEEEtran}
\usepackage{amsmath,amsfonts}
\usepackage{algorithmic}
\usepackage{algorithm}
\usepackage{array}
\usepackage[caption=false,font=normalsize,labelfont=sf,textfont=sf]{subfig}
\usepackage{textcomp}
\usepackage{stfloats}
\usepackage{url}
\usepackage{verbatim}
\usepackage{graphicx}
\usepackage{float}
\usepackage{cite}
\graphicspath{ {img/} }
\usepackage{lettrine}
\usepackage{booktabs}
\usepackage{multirow} 
\usepackage{xcolor} 
\usepackage{colortbl}
\usepackage[colorlinks=true, linkcolor=blue, urlcolor=blue, citecolor=blue]{hyperref}
\begin{document}
\title{Multi-Expert Learning Framework with the State Space Model for Optical and SAR Image Registration}
\author{Wei Wang, Dou Quan,~\IEEEmembership{Member,~IEEE}, Ning Huyan,~\IEEEmembership{Member,~IEEE}, Chonghua Lv,~\IEEEmembership{Student Member,~IEEE}, Shuang Wang,~\IEEEmembership{Senior Member,~IEEE}, Yunan Li,~\IEEEmembership{Member,~IEEE}, Licheng Jiao,~\IEEEmembership{Life Fellow, IEEE}
		\thanks{Manuscript created October 2025, revised December 2025 and January 2026, and accepted February 2026; This work was supported in part by the National Natural Science Foundation of China under Grant 62201407, 62501356, and 62271377; in part by the Key Research and Development Program of Shaanxi Program under Grant 2023QCYLL28, Grant 2024GX-ZDCYL-02-08, and Grant 2024GX-ZDCYL-02-17; in part by the China Postdoctoral Science Foundation under Grant 2022M722496; and in part by the Key Scientific Technological Innovation Research Project by Ministry of Education. (Corresponding author: Dou Quan, Ning Huyan, e-mail: dquan@stu.xidian.edu.cn; quandou@xidian.edu.cn; n-hy@mail.tsinghua.edu.cn.)}
        \thanks{Wei Wang, Dou Quan, Chonghua Lv, Shuang Wang, and Licheng Jiao are with the Key Laboratory of Intelligent Perception and Image Understanding of Ministry of Education of China. Wei Wang is also with the Hangzhou Institute of Technology, Xidian University, Hangzhou 311231, China. Yunan Li is with the School of Computer Science, Xidian University, Xi'an 710071, China. Ning Huyan is with the Department of Automation, Tsinghua University, Beijing 100084, China.}
        
        }
	\markboth{IEEE Transactions on Geoscience and Remote Sensing, October~2025}%
	{Multi-Expert Learning Framework with the State Space Model for Optical and SAR Image Registration}
\maketitle

\begin{abstract}
Optical and Synthetic Aperture Radar (SAR) image registration is crucial for multi-modal image fusion and applications. However, several challenges limit the performance of existing deep learning-based methods in cross-modal image registration: (i) significant nonlinear radiometric variations between optical and SAR images affect the shared feature learning and matching; (ii) limited textures in images hinder discriminative feature extraction; (iii) the local receptive field of Convolutional Neural Networks (CNNs) restricts the learning of contextual information, while the Transformer can capture long-range global features but with high computational complexity. To address these issues, this paper proposes a multi-expert learning framework with the State Space Model (ME-SSM) for optical and SAR image registration. Firstly, to improve the registration performance with limited textures, ME-SSM constructs a multi-expert learning framework to capture shared features from multi-modal images. Specifically, it extracts features from various transformations of the input image and employs a learnable soft router to dynamically fuse these features, thereby enriching feature representations and improving registration performance. Secondly, ME-SSM introduces a state space model, Mamba, for feature extraction, which employs a multi-directional cross-scanning strategy to efficiently capture global contextual relationships with linear complexity. ME-SSM can expand the receptive field, enhance image registration accuracy, and avoid incurring high computational costs. Additionally, ME-SSM uses a multi-level feature aggregation (MFA) module to enhance the multi-scale feature fusion and interaction. Extensive experiments have demonstrated the effectiveness and advantages of our proposed ME-SSM on optical and SAR image registration. Specifically, ME-SSM improves the correct matching rate (CMR) by 7.14\% and 1.95\% based on thresholds 1 and 3, respectively, on the SEN1-2 dataset, and increases the CMR by 2.12\% based on threshold 3 on the OS dataset. The code is available at \href{https://github.com/Miraitowa515/ME-SSM}{https://github.com/Miraitowa515/ME-SSM}.
\end{abstract}

    
\section{Introduction}
\lettrine{O}{ptical} and Synthetic Aperture Radar (SAR) image registration aims to align cross-modal images captured from the same scene \cite{HOPC}. Due to the complementary characteristics of optical and SAR images, they are widely used for multi-modal remote sensing image fusion and various applications. For example, optical images provide rich visual information about the land cover objects, which are easily recognizable and understandable. On the other hand, SAR images can display information about the scattering of surface objects under low-light conditions, or with clouds and fog obscuring. However, the contents of SAR images are challenging to interpret. Therefore, optical and SAR image registration is essential for subsequent multi-modal image fusion and applications. 

\begin{figure}[tbp]  
    \centering
    \includegraphics[width=1\linewidth]{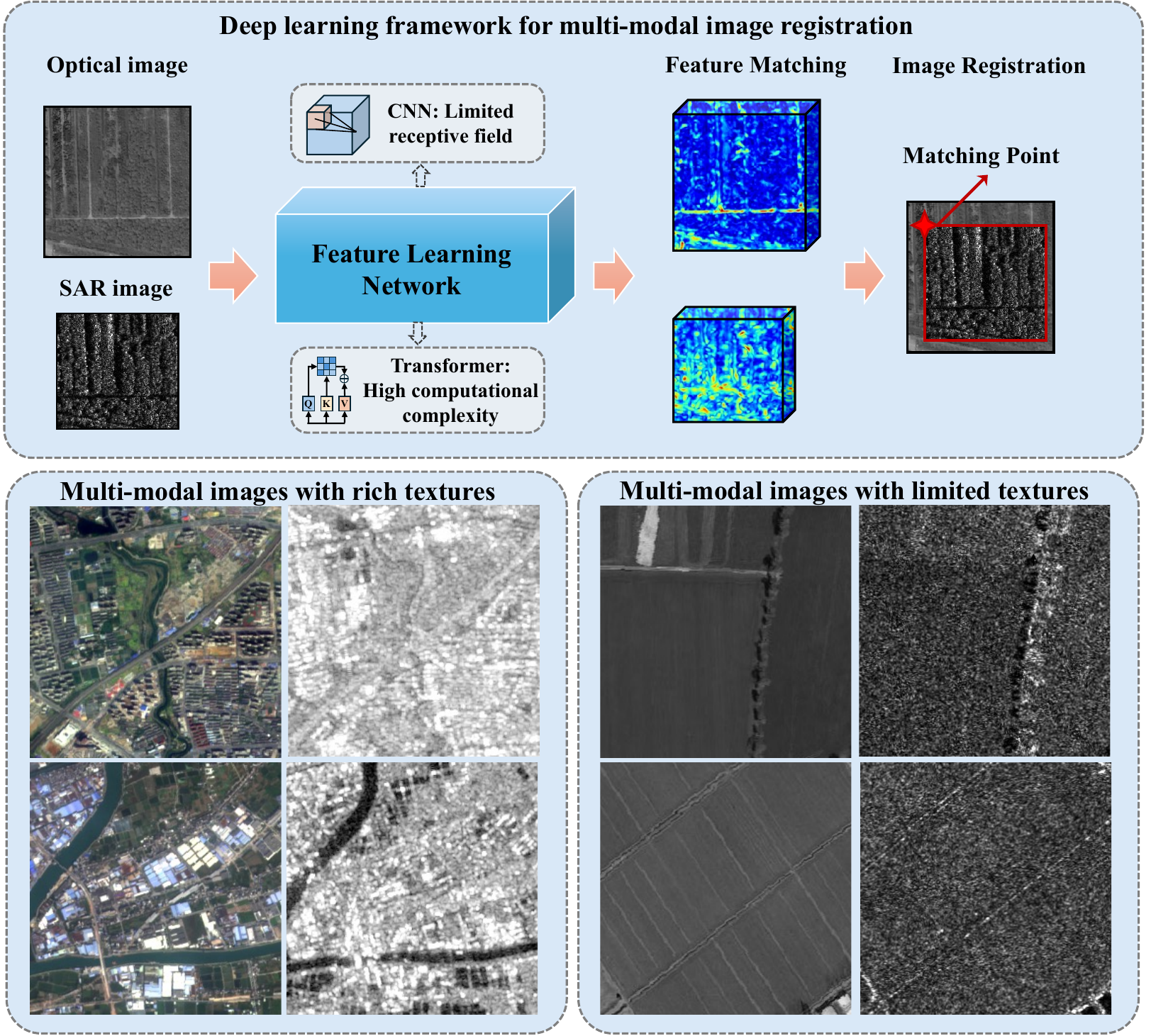}
    \caption{Deep learning frameworks for optical and SAR image registration, which have shown significant performance on images with rich textures, while their performance decreases dramatically when dealing with images with limited textures.}
   \label{multiexpertsintro}
\end{figure}

Deep learning methods have shown significant advantages over traditional region-based methods \cite{FDANet} and handcrafted feature-based methods \cite{SIFT, SURF} in remote sensing image registration. As shown in the upper of Fig. \ref{multiexpertsintro}, deep learning-based methods mainly employ convolutional neural networks (CNNs) \cite{CorrASL, SFcNet, Moganet} to learn shared features from cross-modal images, and then build local correspondences across images for registration by feature matching or align images according to global features by template matching. Existing deep learning-based methods try to boost registration performance from the following aspects, such as using dilated convolutional operators to expand receptive fields \cite{dilated_convolutional}, employing multi-scale network architectures to preserve more detailed information \cite{SiameseUNet/1, MARU-Net}, adopting a two-stage matching framework to perform coarse-to-fine registration \cite{RSOMNet}, utilizing frequency domain information from multi-modal images to enhance feature discriminability \cite{F3Net}, combining handcrafted features with deep features for image registration \cite{AESF}, and so on. 

However, current deep learning methods for optical and SAR image registration still have several limitations. Firstly, deep learning methods have shown significant performance on remote sensing image scenes with rich texture features, while their performance drops sharply when dealing with images that have limited texture features. As shown in the bottom of Fig. \ref{multiexpertsintro}, the multi-modal images in the right part have fewer textures than those in the left part. The main reason is that a single deep model is difficult to extract discriminative features from these images for registration. Secondly, due to local receptive fields, CNNs cannot capture global information, which restricts their ability to achieve optimal registration performance. Although transformer structures can capture global features from images, they will bring expensive computational costs.

To solve these problems, this paper proposes a multi-expert learning framework with the State Space Model (ME-SSM) for optical and SAR image registration. Inspired by Mixture-of-Experts (MoE) \cite{ConvMoE/1, multimodal_MOE}, ME-SSM builds a multi-expert learning framework to capture shared features from cross-modal images for further improving image registration performance with limited texture features. Specifically, ME-SSM applies different affine transformations to the input images and extracts salient features using multiple experts from the transformed image features. After that, ME-SSM dynamically aggregates these features through a learnable soft router, capturing rich and discriminative features and significantly boosting registration performance. Additionally, ME-SSM first introduces a State Space Model, Mamba \cite{Mamba}, into remote sensing image registration. ME-SSM can model long-range global information with linear complexity through multi-directional cross-scanning strategies, effectively capturing global contextual features from remote sensing images with a small computational cost for accurate image registration. Moreover, ME-SSM uses a multi-level feature aggregation (MFA) module in the Mamba to enhance the multi-scale feature fusion and interaction, thereby improving image registration accuracy. It should be noted that the proposed multi-expert learning framework achieves performance gains under various feature extraction network architectures, such as CNNs, ViTs, and Mamba.

The main contributions of this paper are as follows:

\begin{itemize}
    \item This paper proposes a multi-expert learning framework with the State Space Model (ME-SSM) for optical and SAR image registration. ME-SSM can capture rich global contextual information for more accurate cross-modal image registration.

    \item This paper designs a general multi-expert learning framework (MELF) for enhancing the feature learning from images with limited textures. It adopts multi-expert feature learning and dynamic aggregation based on the various affine transformations of the input image. MELF can be easily and quickly integrated into different frameworks (i.e., CNNs, ViTs, and Mamba) and yields significant performance improvements.

    \item ME-SSM first introduces the state space model for remote sensing image registration, which can capture global features for facilitating registration accuracy. 

    \item Experimental results on public optical and SAR image registration datasets with various image resolutions,  SEN1-2 and OS datasets, have shown the effectiveness and advantages of the proposed ME-SSM. 
\end{itemize}

\section{Related Works}

\subsection{Remote Sensing Image Registration}

Research on remote sensing image registration generally falls into two mainstream categories: mapping-based image registration and pixel-shift matching. Firstly, mapping-based image registration \cite{mapping_1,mapping_2} methods formulate image registration as a geometric parameter estimation problem for image transformation and alignment. They aim to estimate an explicit geometric mapping (e.g., affine or homography transformation) between two images and perform image alignment in the same spatial coordinate system by global image transformation. The mapping-based methods usually directly predict the transformation parameters or estimate the transformation parameters based on the local matching points or correspondences. In contrast, pixel-shift matching methods formulate the image registration task as a template matching problem, finding the matching position of the template image within the reference image. Template matching methods extract the global features from the template and reference images for similarity map computation and obtain the optimal matching position by identifying the peak value of the similarity map. In practical applications, images can be coarsely aligned using sensor models and auxiliary information. To adopt the global features for more precise image registration, this paper primarily focuses on the fine-grained image registration based on the pixel-shift matching method. Existing methods in this field have evolved from traditional methods to deep learning-based methods.

Traditional methods can be categorized into feature-based methods and region-based methods. Feature-based methods aim to extract invariant and discriminative features to establish local correspondences between images based on feature similarity. Traditional handcrafted features are related to gradient statistical information or shape features, such as SIFT\cite{SIFT}, SURF\cite{SURF}, SAR-SIFT\cite{SAR_SIFT}, and Oriented FAST and Rotated BRIEF (ORB) \cite{ORB}. However, these methods cannot extract consistent features from optical and SAR images, which are sensitive to image modality changes. Region-based methods, also known as template matching methods, involve sliding a small template pixel by pixel over the reference image and searching for matching parameters by maximizing the image similarity metric. The common similarity metrics include Sum of Squared Differences (SSD) \cite{SSD_NCC}, Normalized Cross-Correlation (NCC) \cite{NCC}, and Mutual Information (MI) \cite{MI/1}. However, traditional template matching methods calculate image similarity directly based on grayscale information. Thus, they are sensitive to radiometric differences between multi-modal images. To mitigate non-linear radiometric differences, structural information \cite{HOPC} and multi-directional pixel gradient features \cite{CFOG, AWOG} are proposed to enhance the discriminative and robustness features for improving registration accuracy. However, traditional handcrafted descriptors mainly rely on low-level statistical information, which fail to extract shared and discriminative representations from multi-modal images and are susceptible to severe speckle noise.

Deep learning-based methods are gradually replacing handcrafted methods and bringing significant improvements for remote sensing image registration. Deep learning-based methods mainly employ the deep convolutional network to extract local feature descriptors from image patches surrounding keypoints or to learn global feature maps from whole images \cite{HardNet, Matchnet}. Compared with handcrafted features, deep features have stronger invariance and discrimination, boosting the image registration accuracy. Existing deep learning methods mainly improve remote sensing image registration performance from several aspects: Siamese networks with U-Net architectures \cite{SiameseUNet/1, VUNet}, deep phase features learning \cite{MOPSI, PCNet}, attention-enhanced structural feature learning \cite{AESF}, multi-scale feature learning \cite{Multiscale2022}, dense feature representation learning \cite{RMSO-ConvNeXt}, frequency feature learning \cite{FFTUnet, DWNet}, and lightweight efficient matching \cite{LM-Net}. 

However, cross-modal image registration still faces several challenges. The limited receptive field of CNNs restricts their ability to capture spatial contextual information for image registration. Although transformer-based structures can extract global features, they have expensive computational costs. Additionally, existing approaches struggle to extract discriminative features from remote sensing images with poor texture. To address these challenges, this paper proposes a multi-expert learning framework with the State Space Model, ME-SSM, for optical and SAR image registration. It employs multiple experts to extract features from variously transformed images, enhancing the richness and discriminative power of features under low-texture conditions. Meanwhile, ME-SSM leverages a state space model to enhance global feature learning with linear complexity, boosting image registration efficiency and performance.

\subsection{Mixture of Experts}
Mixture of Experts (MoE) method \cite{MOE/1} usually decomposes the task into multiple sub-tasks and employs a set of specialized models, experts, to address these sub-tasks. MoE can improve both performance and efficiency, and is widely used in natural language processing and computer vision. MoE consists of multiple identical expert models and a trainable gating network, router. The common expert model utilizes Feed-Forward Networks (FFN) \cite{FFN}, which can be integrated into Transformer blocks, providing computational advantages through parameter sparsity and task specialization. In multi-modal image processing tasks, the modality-specific convolutional network or transformer encoder can be viewed as experts to deal with images with different modalities \cite{multimodal_MOE}. The router dynamically scores each expert and selects the appropriate one using a gating network \cite{router/1, router/2}. The gating networks contain several types. For example, dense gating \cite{DenseGate} activates all experts simultaneously and combines their outputs. Sparse gating \cite{sparseGate} selects one or a few experts based on their gating scores, reducing computational costs during large-scale model training. Soft gating \cite{SoftGating} aggregates the outputs of experts using weighted sums, improving model performance in fine-grained scenarios. Token-choice gating \cite{Token-Choice_Gating} dynamically selects the suitable expert according to the input token. 

Inspired by the effective and diverse learning abilities of each expert in MoE, this paper builds a multi-expert learning framework to extract rich features for improving the performance of optical and SAR image registration.

\subsection{State Space Models}
In recent years, the state space model (SSM) \cite{SSM/1, SSM/2} has gained significant attention due to its global feature learning ability with linear complexity. Inspired by the SSM, Gu et al. \cite{Mamba} introduce Mamba, a deep learning architecture that combines the recursive properties of recurrent neural networks (RNNs) with the parallel computation capabilities of Transformers, which achieves superior performance in natural language processing tasks. Compared to the Transformer, Mamba can handle longer input sequences with lower computational complexity. 
Vision Mamba \cite{Vision_mamba} extends the Mamba architecture for computer vision tasks and builds a vision backbone with bidirectional Mamba blocks, achieving significant advantages in computation efficiency.

Due to the advantages of linear complexity and global receptive fields, SSMs have been introduced into remote sensing image perception and interpretation tasks. For example, RSMamba \cite{chen2024rsmamba} adopts the SSM for remote sensing image classification. MaDiNet \cite{MaDiNet} proposes a Mamba with a diffusion model for SAR target detection, which can capture spatial structural information and enhance the target detection. MBSSNet \cite{MBSSNet} proposes a joint semantic segmentation network based on Mamba for optical and SAR images, which utilizes the multi-scale and multi-modal fusion to mine spatial and semantic information for enhancing segmentation accuracy. Pan et al. \cite{M3-CR} use the multi-scale multibranch Mamba model for optical image cloud removal, which can model long-range global dependencies in images and perform deep feature interaction and integration between optical and SAR images, thereby boosting SAR-assisted thick cloud removal. Moreover, Zhao et al. \cite{HSFMamba} propose a hierarchical selective fusion Mamba network for SAR image super-resolution and denoising, which can effectively integrate optical priors by cross-selection scan mechanisms to facilitate SAR high-quality reconstruction.

Although SSMs have demonstrated significant potential in high-level remote sensing image processing tasks, their application in low-level vision tasks, particularly in optical and SAR image registration, has not yet been fully explored. In this paper, we introduce SSM for optical and SAR image registration. By employing a multi-directional cross-scanning strategy, the deep model can capture long-range global contextual information with linear computational complexity. Additionally, we integrate the multi-level feature aggregation module in the feature extraction process, enhancing the multi-scale feature fusion and interaction, and thereby improving image registration accuracy.

\begin{figure*}[t] 
    \centering
    \includegraphics[width=\textwidth]{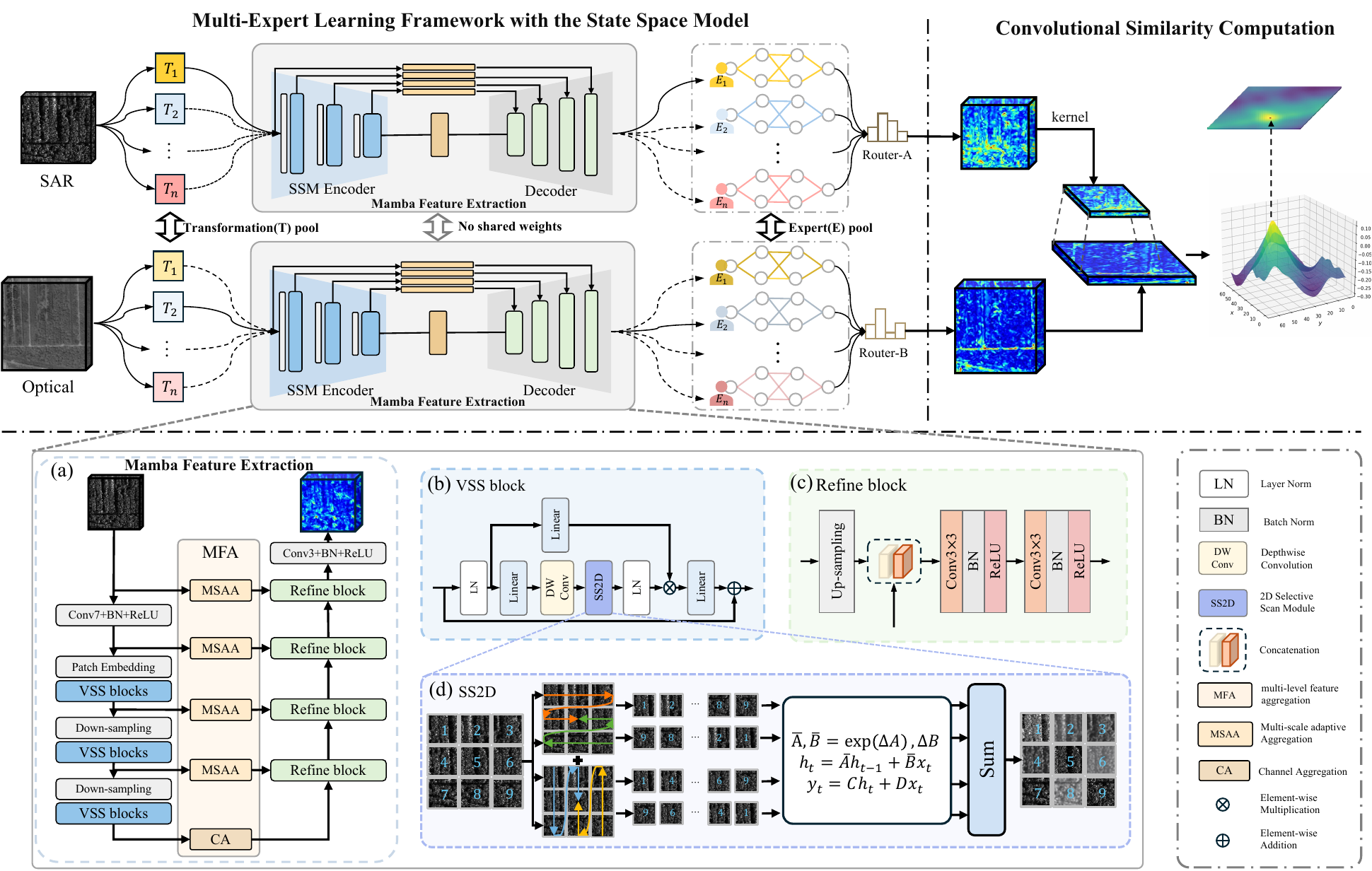} 
    \caption{The pipeline of the proposed multi-expert learning framework with the State Space Model (ME-SSM) for optical and SAR image registration. ME-SSM utilizes a multi-expert learning framework to dynamically aggregate rich features from different transformed images by various experts and a learnable soft router. ME-SSM employs the State Space Model, Mamba, for feature extraction, which can effectively capture global spatial information and multi-level features. Ultimately, we utilize the normalized SAR feature as a convolution kernel and perform a sliding convolution on the optical feature for fast similarity calculation, thereby achieving the optical and  SAR image registration.}
    \label{fig:pipeline}
\end{figure*}

\section{Preliminaries}
State space models (SSMs) are mathematical modeling approaches that center around the system's state and describe the behavior and evolution of dynamic systems over time. It is widely applied in control systems, time series analysis, and signal processing. Mamba and a series of models based on SSM are inspired by linear time-invariant systems, which describe the continuous dynamics of a system using ordinary differential equations (ODEs). These ODEs map the input signal to a latent space. Mathematically, it can be expressed as:
\begin{equation}
\left\{\begin{array}{l}
h^{\prime}(t)=\mathbf{A} h(t)+\mathbf{B} x(t), \\
y(t)=\mathbf{C} h(t)+\mathbf{D} x(t),
\end{array}\right.
\end{equation}
where $x(t)\in R$ represents the input sequence signal, $y(t)\in R$ represents the output signal, and $h(t)\in \mathbb{R}^{N}$ represents the latent state. $h'(t)\in \mathbb{R}^{N}$ represents the time derivative of $h(t)$. Additionally, $A\in \mathbb{R}^{ N\times N}$ serves as the state transition matrix, with $B\in \mathbb{R}^{ N\times 1}$,$C\in \mathbb{R}^{ 1\times N}$ being projection matrices, while $D\in \mathbb{R}^{1}$ commonly acts as a residual connection. Specifically, the parameter $A$ stores historical information, and $B$ controls the influence of the current input $x(t)$ on the hidden state. The output $y$ is generated by both the hidden state and the input $x(t)$.

In Mamba\cite{Mamba}, the ordinary differential equations are discretized using a zero-order hold, and the continuous parameters $A$ and $B$ are transformed into discrete parameters $\bar{A}$ and $\bar{B}$ via a time-scale parameter $\bigtriangleup$, which can be expressed as:

\begin{equation}
\left\{\begin{array}{l} \mathbf{\bar{A}}=exp(\bigtriangleup \mathbf{A}), \\ \mathbf{\bar{B}}=(\bigtriangleup \mathbf{A})^{-1}(exp(\bigtriangleup \mathbf{A})-I) \cdot \bigtriangleup \mathbf{B}. \end{array}\right.
\end{equation}

After discretization, the system can be expressed as:

\begin{equation}
\left\{\begin{array}{l} h'_{t}=\mathbf{\bar A} h_{t-1}+\mathbf{\bar B} x_{t}, \\ y_{t}=\mathbf{C} h_{t}. \end{array}\right.
\end{equation}
To enhance computational efficiency, the sequence $x$  with length $N$ can be computed using a global convolution to produce the output $y$. The specific equation can be expressed as:

\begin{equation}
\left\{\begin{array}{l} \mathbf{\bar{K}}=(\bar{\mathbf{C}}\bar{\mathbf{B}},\bar{\mathbf{C}}\bar{\mathbf{A}}\bar{\mathbf{B}},\bar{\mathbf{C}}\bar{\mathbf{A^{2}}}\bar{\mathbf{B}},\cdots,\bar{\mathbf{C}}\bar{\mathbf{A}}^{N-1}\bar{\mathbf{B}}),  \\ y=x\ast \mathbf{\bar{K}},
\end{array}\right.
\end{equation}
where $\ast$ denotes the convolution operation, and $\mathbf{\bar{K}}$ represents the convolution kernel. However, the traditional SSM is linear and time-invariant, meaning that the projection matrices do not change with the input. This limitation prevents the model from considering the importance of different time steps. Mamba addresses this by using a selective scanning mechanism, mapping the parameter matrices as:

\begin{equation}
\left\{\begin{array}{l} \bar{\mathbf{B}}=s_{B}(x), \\ \bar{\mathbf{C}}=s_{C}(x), \\ \bigtriangleup =\tau _{\bigtriangleup } (Parameter+s_{\bigtriangleup}(x)). \end{array}\right.
\end{equation}
where $Parameter$ is a randomly selected set of values and the model dynamically adjusts its parameters based on the input. It transforms the SSM into a linear time-varying system, thereby improving its performance in processing complex sequences.

\section{Methodology}

\begin{figure}[t] 
    \centering
    \includegraphics[width=\linewidth]{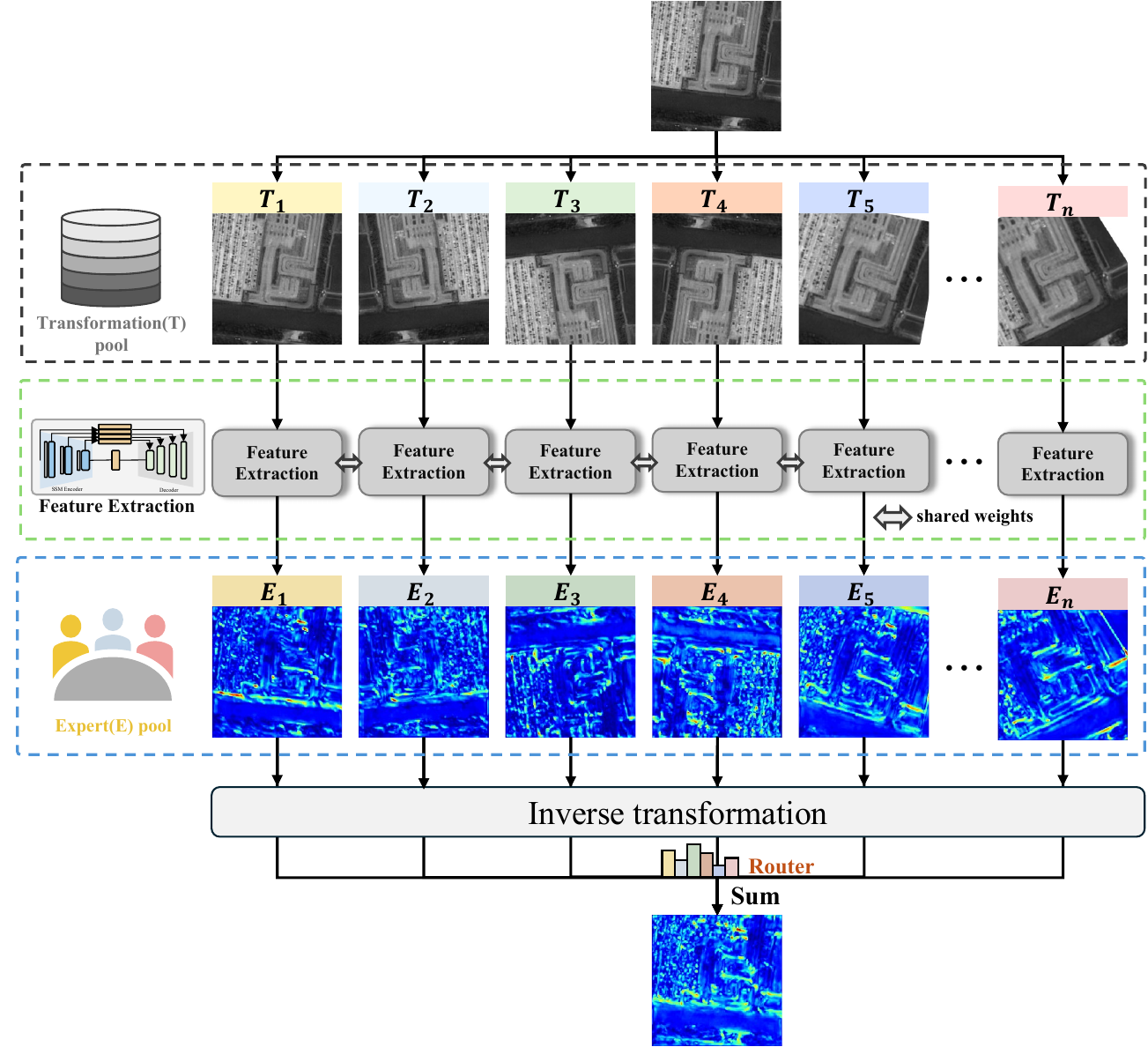} 
    \caption{The pipeline of the multi-expert learning framework. }
    \label{fig:moe_router}
\end{figure}

\subsection{Problem Formulation}
Given a template SAR image, $X_{S}$, and a reference optical image, $X_{O}$, cross-modal image registration aims to locate the matching position of the template in the reference image. The core of cross-modal image registration is to extract invariant features to the image modality change and discriminative features to distinguish the non-matching location from optical and SAR images (i.e., $F_{S}$, $F_{O}$), enabling the feature similarity of optical and SAR images in the matching position to be larger than that in non-matching positions.

\subsection{Overall Framework}
Figure \ref{fig:pipeline} illustrates the pipeline of the multi-expert learning framework with the State Space Model (ME-SSM) for optical and SAR image registration. ME-SSM adopts multiple experts to extract features from the geometrically transformed variants of the input images and dynamically aggregates the learned features of multiple experts, enhancing discriminative feature learning and boosting image registration performance. Additionally, ME-SSM introduces a State Space Model, Mamba, into the feature learning process, further enhancing the feature representation. Finally, we register optical and SAR images by computing the similarity of their extracted features. 

\subsection{Multi-expert Learning Framework}
As shown in Fig. \ref{fig:moe_router}, this paper proposes a multi-expert learning framework to enhance feature learning by aggregating the features of multiple experts, where each expert extracts features from a geometrically transformed variant of the input image. Specifically, we design an image transformation pool and a corresponding expert pool. In the image transformation pool, we apply random geometrical transformations to input images, including flipping, random rotation, and homography transformations. Then, we employ multiple experts to extract features from these transformed images. The selected image transformations should preserve the textural information in images, which can avoid the smoothing effect caused by bilinear or bicubic interpolation in large affine transformations. Meanwhile, the image transformations are reversible, ensuring pixel-level alignment of the features for subsequent feature fusion. Specifically, we adopt identity, horizontal flip, vertical flip, and central rotation in experiments.

The multi-expert learning process can be formulated as follows:
\begin{align}
    &X_{S,O}^{i} = T_{i}(X_{S,O}),\\
    &F_{S,O}^{i} = F(X_{S,O}^{i}),\\
    &\bar{F}_{S,O}^{i} = E_{i}(F_{S,O}^{i}),
\end{align}
where $X_{S,O}$ represents the input image $X_{S}$ or $X_{O}$, $T_i$ represents the $i_{th}$ image transformation, $F$ is the shared feature learning model for various transformed images, $E_i$ denotes the $i_{th}$ expert, $\bar{F}_{S,O}^{i}$ refers to the features extracted by the $i_{th}$ expert.

To maintain computational efficiency, each expert $E_i$ employs a lightweight architecture. Specifically, it consists of only a single standard $1\times1$ convolutional layer. We omit normalization (e.g., batch normalization) and nonlinear activation functions (e.g., ReLU) to preserve the intrinsic feature distribution of the transformed images. Each expert focuses on channel-wise weighting and distribution alignment, thereby enhancing features from transformed images with distinct geometric perspectives. Multiple experts can fuse these features to boost image registration performance.

After that, the extracted features of different transformed images are recalibrated through the corresponding inverse transformations. Unlike traditional MoE \cite{sparseGate} methods that sparsely select experts for fusion, we adaptively fuse the features of all experts through a soft router. The soft router is implemented as a global learnable weighting mechanism and optimized by an end-to-end learning mode. The proposed multi-expert learning can enrich the feature representation and enhance the discrimination of features, improving the image registration performance. The details of this process can be represented as follows:
\begin{align}
    &\hat{F}_{S,O}^{i} = IT_{i}(\bar{F}_{S,O}^{i}),\\
    &F_{S,O} = {\textstyle \sum_{i=0}^{n-1}}  w_{i}\cdot \hat{F}_{S,O}^{i},\\
    &w_{i} = \delta(\alpha_{i}),
\end{align}
where $IT_i$ represents the inverse transformation corresponding to the $i_{th}$ transformed image, $\delta$ is the softmax function, $w_{i}$ is the fusion weight of the features of the $i_{th}$ expert, $\alpha_{i}$ represents the learnable soft router parameter of the $i_{th}$ expert, and $F_{S,O}$ is the aggregated feature from various experts.

It should be noted that the proposed multi-expert learning strategy has strong generalizability and can be seamlessly integrated into various deep network frameworks, significantly improving the performance of cross-modal image registration.

\subsection{Mamba for Feature Learning}
In this paper, we introduce the state space model, Mamba, in the feature extraction process, which can enhance global feature learning. As shown in Fig. \ref{fig:pipeline}(a), the feature extraction can be divided into three stages: global context feature extraction, multi-level feature aggregation, and feature interaction and refinement. In the first stage, ME-SSM integrates visual state space (VSS) blocks \cite{VMamba} into the feature extraction, which employs a multi-directional cross-scanning strategy based on a state space model to capture spatial global contextual information from images. In the second stage, ME-SSM utilizes multi-scale adaptive aggregation and channel aggregation to enhance local feature learning. In the third stage, ME-SSM performs feature interaction and refinement to fuse the global and local features and improve feature discrimination.

\subsubsection{Global context feature extraction}
Taking an input image $X \in \mathbb{R}^{H \times W \times 1}$ as an example, we first use a convolutional block to extract features from the input image. The convolutional block contains a convolutional layer with a kernel size of $7 \times 7$, batch normalization, and a ReLU function. After that, we segment the extracted feature map $F_0$ into patches and project them into a sequence input $F'_{0} \in \mathbb{R}^{ \frac{H'}{P} \times \frac{W'}{P} \times D}$, where $P$ represents the patch size and $D$ represents the latent dimension. Next, we adopt several VSS blocks to model the spatial contextual information and achieve hierarchical feature representations $F'_{i}$, where $i \in \{1, 2, 3\}$. 

The structure of the VSS block is presented in Fig. \ref{fig:pipeline}(b). The input feature is first processed through layer normalization and then split into two branches. One branch contains a linear embedding layer, followed by a depth-wise convolution (DWConv) layer, a 2D selective scanning module (SS2D) for contextual modeling, and a layer normalization. The other branch contains a linear embedding layer. The outputs of these two branches are fused with the input feature by a residual connection to produce the final output. 

In SS2D, we encode the image patch sequences from four different directions, creating four distinct sequences. Each sequence is processed using the SSM, and the final feature map is obtained by merging the output features from the four sequences. As shown in Fig. \ref{fig:pipeline}(d), for the input feature $f$, the sequence process can be expressed as follows:
\begin{align}
&f_{j}=Expand(f,j),\\
&\bar f_{j} = S6(f_{j}),\\
&F'=Merge(\bar f_{1}, \bar f_{2}, \bar f_{3} ,\bar f_{4}),
\end{align}
where $Expand(\cdot)$ denotes the scanning expansion (i.e., creating a sequence based on a scanning direction), $j \in \{1, 2, 3, 4\}$ corresponds to the four different scanning directions, the selective cross-scanning spatial state sequential model $S6$  serves as the core operator, enabling each pixel in the image to effectively integrate information from all pixels across different directions. $Merge(\cdot)$ means the scanning merging process (i.e., adding the sequences processed by the $S6$ module from different scanning directions and reshaping the result into a 2D feature map).

\subsubsection{Multi-level feature aggregation}
We utilize the multi-level feature aggregation (MFA) module to further enhance feature representation at each level of the SSM encoder. It comprises two parts: multi-scale adaptive aggregation (MSAA) and channel aggregation (CA).

As shown in Fig. \ref{fig:MFA_module}(a), the MSAA module fuses multi-scale features based on a series of depth-wise convolutions at various scales. For the input feature $ F'_{i}$ , the details of MSAA are as follows:
\begin{figure}[tbp]
    \centering
    \begin{minipage}{\linewidth}
        \centering
        \includegraphics[width=0.9\linewidth]{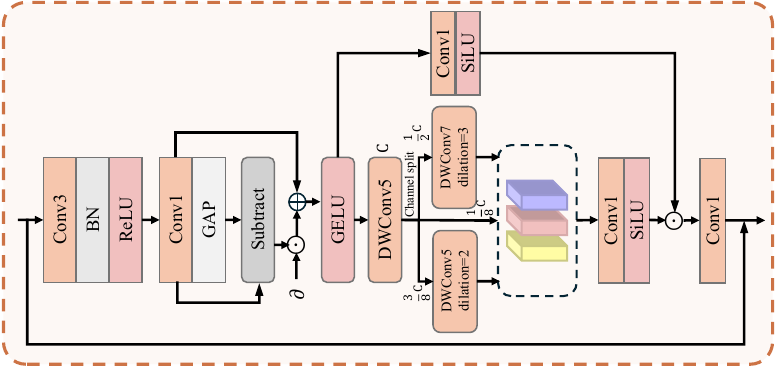} 
        \par {\footnotesize (a) MSAA module}
        \label{fig:module1}
    \end{minipage}
    \vspace{2pt} 
    \begin{minipage}{\linewidth}
        \centering
        \includegraphics[width=0.9\linewidth]{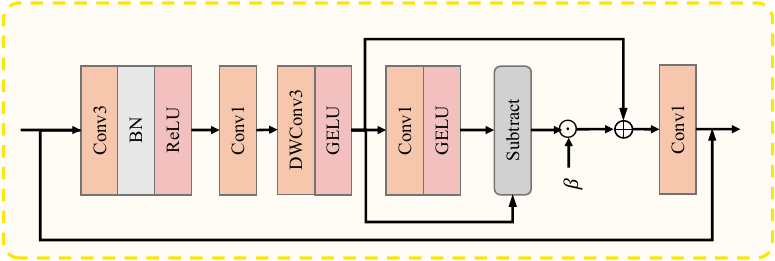} 
        \par  {\footnotesize (b) CA module}
        \label{fig:module2}
    \end{minipage}
    \vspace{2pt}
    \begin{minipage}{\linewidth}
        \centering
        \includegraphics[width=0.9\linewidth]{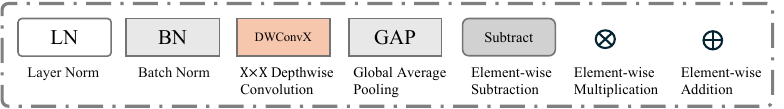} 
        \label{fig:module3}
    \end{minipage}
    \caption{The multi-level feature aggregation (MFA) module comprises two parts: multi-scale adaptive aggregation (MSAA) and channel aggregation (CA), which can further enhance feature representation.}
    \label{fig:MFA_module}
\end{figure}

\begin{equation}
    \left\{
    \begin{array}{l}
    Y_i = \operatorname{Conv}_{1}(\text{ReLU}(\text{BN}(\operatorname{Conv}_{3}(F'_{i})))), \\
    Y'_i = \operatorname{GELU}\left(Y_i + \alpha \odot \left(Y_i - \operatorname{GAP}(Y_i)\right)\right), \\
    \bar{Y_i} = \operatorname{DWConv}_{5}(Y'_i), \\
    Z'_i = \delta_{1} \operatorname{DWConv}_{5}(\bar{Y_i}) \oplus_c \delta_{2} \operatorname{DWConv}_{7}(\bar{Y_i}) \oplus_c \delta_{3} \bar{Y_i}\\
    \bar{Z_i} = \operatorname{SiLU}\left(\operatorname{Conv}_{1}(Z'_i)\right) \odot \operatorname{SiLU}\left(\operatorname{Conv}_{1}(Y'_i)\right),\\
    Z_i = F'_{i} + \operatorname{Conv}_{1}(\bar{Z_i}),
    \end{array}
    \right.
\end{equation}
where  $\alpha$  denotes the scaling factor, $\delta_{1}$, $\delta_{2}$, and $\delta_{3}$ denote the proportions used to partition the features along the channel dimension, which are $\frac{3}{8}, \frac{1}{2}, \frac{1}{8}$, respectively, $\oplus_c$ denotes the concatenation operation, $\operatorname{Conv}_{l}$ represents a convolution with a kernel size of $l \times l$, $\operatorname{DWConv}_{l}$ represents a depthwise convolution with a kernel size of $l \times l$, and $SiLU$ is the Sigmoid Linear Unit function.

Meanwhile, we employ a CA module in the high-level features $F'_{3}$ to improve both channel efficiency and feature representation, all while maintaining a small increase in network parameters and computational cost. The details are as follows:
\begin{equation}
    \left\{
    \begin{array}{l}
    Y_3 = \operatorname{Conv}_{1}(\text{ReLU}(\text{BN}(\operatorname{Conv}_{3}(F'_{3})))), \\
    \bar{Y}_3 = \text{GELU}(\operatorname{DWConv}_{3}(Y_3)), \\
    Z'_3 = \bar{Y}_3 + (\bar{Y}_3 - \beta \odot (\text{GELU}(\operatorname{Conv}_{1}(\bar{Y}_3)))), \\
    Z_3 = F'_{3} + \operatorname{Conv}_{1}(Z'_3),
    \end{array}
    \right.
\end{equation}
where $\beta$ denotes the scaling factor. 

\subsubsection{Feature interaction and refinement}
After acquiring multi-level features, we employ the feature refinement layers and upsampling for adaptive fusion. Finally, the fused features can be used for similarity computation and cross-modal registration.
\begin{equation}
    R_{j}=Fuse(UP(R_{j-1}),{Z}_j).
    \label{eq:third_stage}
\end{equation}
where $UP$ represents the upsampling operation, $Fuse$ refers to the feature channel concatenation and adaptive fusion operation in the feature refinement layer, $R_{j}$ and $R_{j-1}$ represent the fused features at the current layer and the previous layer, respectively, and ${Z}_j$ denotes the output features of the MFA module at the current layer.

\begin{figure}[htbp] 
    \centering
    \includegraphics[width=\linewidth]{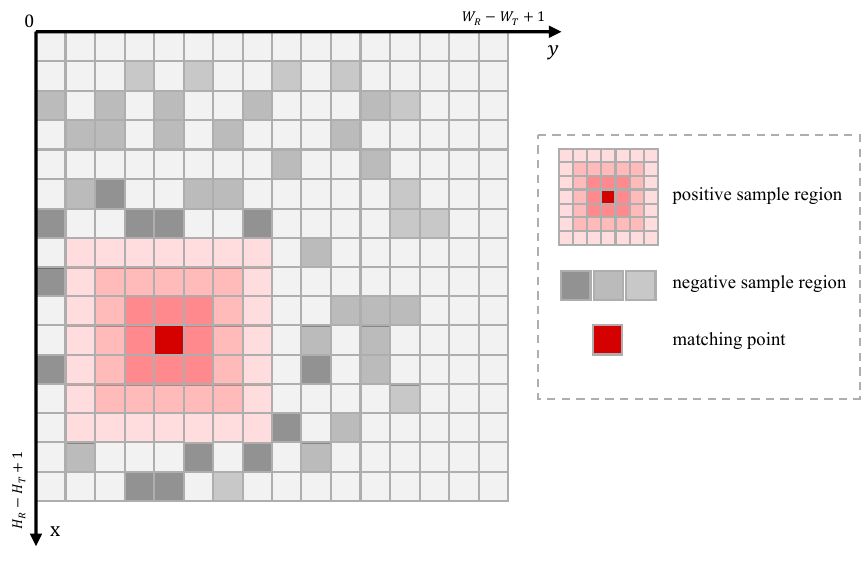} 
    \caption{The similarity map $S$. The positive sample region is a soft-label region obtained by applying Gaussian smoothing centered on the matching point. 
    The negative samples used for deep model optimization are selected from the $top k$ similarity candidates that fall outside of the positive sample region.}

    \label{fig:similarity_grid}
\end{figure}

\subsection{Similarity Computation and Loss Function}
\subsubsection{Similarity computation}
Similarity calculation between the template and the reference image through pixel-by-pixel comparison is highly time-consuming, significantly increasing training and inference times. To accelerate this process, we employ a convolution-based approach for similarity computation. Specifically, we take the feature map of the template image as a convolution kernel and perform convolution on the feature map of the reference image.

The similarity computation can be represented as follows:
\begin{equation}
S[i,j] = \frac{\sum_{c=1}^{C} \sum_{k=1}^{H_{S}} \sum_{l=1}^{W_{S}} F_{S}[c, k, l] \cdot F_{O}[c, i+k, j+l]}{\|{F_{S}}\|_{2} \cdot \|{F_{O}}\|_{2}},
\end{equation}
where $S[i,j]$ represents the feature similarity of the template image $X_{S}$ at position $(i, j)$ in the reference image $X_{O}$, $F_{S} \in \mathbb{R}^{C \times H_{S} \times W_{S}}$ denotes the feature map of the template image, $F_{O} \in \mathbb{R}^{C \times H_{O} \times W_{O}}$ denotes the feature map of the reference image, and the size of the similarity map is $(H_{O} - H_{S} + 1) \times (W_{O} - W_{S} + 1)$. 

\subsubsection{Loss Function}
Following the previous cross-modal image registration \cite{F3Net}, this paper employs three loss functions for network optimization: the matching loss, $L_{m}$, the fine similarity loss, $L_{fine}$, and the similarity peak loss, $L_{peak}$. 

\textbf{Matching loss:} As illustrated in Fig. \ref{fig:similarity_grid}, the matching point and a small surrounding region are positive samples, and other samples are negative samples. The matching loss aims to maximize the similarity of the positive samples and minimize the similarity of the negative samples. The matching loss can be represented as follows: 
\begin{equation}
    L_{m}=(S_{neg}+1)^{2} + (1-S_{pos})^{2},
\end{equation}
where $S_{pos}$ and $S_{neg}$ represent the similarity of positive and negative samples, respectively. In this work, we select a $7 \times 7$ region that surrounds the ground truth as the positive sample. Additionally, we select negative samples with the $top k$ similarity for boosting network optimization, and $k = 49$. 

\textbf{Fine similarity loss:} To further enhance the network optimization and find the accurate matching location, this paper uses the fine similarity loss on the positive samples for network training. It constrains the similarity values to decrease gradually as the distance from the ground truth position increases. It can be formulated as follows:
\begin{equation}
    \begin{array}{l} 
    G_{gt}=Gaussian(gt), \\ 
    L_{fine}=(topk'(G_{gt})-S_{pos}^{topk'})^{2},
    \end{array}
\end{equation}
where  $G_{gt}$ represents the soft label region obtained by applying Gaussian smoothing to the ground truth location.  $S_{pos}^{topk'}$ denotes the subset of positive sample region corresponding to the $top k'$ positions of $G_{gt}$ and $k'=9$.

\textbf{Similarity peak loss:} Additionally, this paper uses the similarity peak loss to constrain the similarity map to have a single peak at the matching location. It can be formulated as follows:
\begin{equation}
    L_{peak} = 2-(max(S)-mean(S)),
\end{equation}
where $max(\cdot)$ and $mean(\cdot)$  denote maximum and average operators, respectively.

The overall loss function for network optimization can be expressed as:
\begin{equation}                        
L_{final}=L_{m}+\gamma_{1}L_{fine}+\gamma_{2} L_{peak},
\end{equation}
where $\gamma_{1}$ and $\gamma_{2}$ denote the hyper-parameters corresponding to the losses, respectively.

The feature extraction network, experts, and soft router are simultaneously trained by an end-to-end learning mode. This training strategy enables the deep model to dynamically and adaptively fuse the features from various geometric views and provide more robust features for optical and SAR image registration.

\section{Experiments and results}

In this section, we evaluate the effectiveness and advantages of the proposed ME-SSM for optical and SAR image registration. Firstly, we introduce the datasets and experimental details. Secondly, we compare ME-SSM with several existing representative methods on image registration. Subsequently, we conduct ablation studies and analyze the effectiveness of the multi-expert learning framework and multi-level feature aggregation. Finally, we present the visualization results and display the robustness of our proposal to noise.

\begin{table}[t]
    \caption{Comparison results of the average $L_{2}$ distance and CMR on the SEN1-2 dataset. The top two results are marked as red and blue, respectively.}
    \centering
    \resizebox{\linewidth}{!}{
    \begin{tabular}{@{}lcccccc@{}}
    \toprule
                              &                               & \multicolumn{4}{c}{CMR(T) (\%) $\uparrow$}                                                                                                                              \\ \cmidrule(l){3-6} 
    \multirow{-2}{*}{Methods} & \multirow{-2}{*}{$L_{2}$ $\downarrow$} & $T=1$                          & $T=2$                          & $T=3$                          & $T=5$                          \\ \cmidrule(r){1-6}
    NCC\cite{Standard_NCC}              & 31.60                          & 5.00                          & 11.00                         & 16.00                         & 23.00                         \\
    CFOG\cite{CFOG}                      & -                             & 26.46                        & 40.67                        & 49.79                        & -                            \\
    DDFN\cite{DDFN}                      & 17.50                          & 30.00                         & 42.00                         & 50.00                         & 55.00                         \\
    MI\cite{MI}                        & 11.80                          & 30.00                         & 45.00                         & 54.00                         & 62.00                         \\
    Siamese CNN\cite{dilated_convolutional}               & 8.27                          & 42.00                         & 61.00                         & 72.00                         & 78.00                         \\
    FFT U-Net\cite{FFTUnet}                 & 6.92                          & 44.00                         & 63.00                         & 74.00                         & 80.00                         \\
    MARU-Net\cite{MARU-Net}                  & 4.94                          &  55.00  & 75.00                         & 82.00                         & 87.00                         \\
    F3Net\cite{F3Net}                    &  4.15                        & 44.72                        & {\color[HTML]{3531FF} 81.03} & {\color[HTML]{3531FF} 87.24} &  89.97 \\
    
    VU-Net\cite{VUNet} & {\color[HTML]{3531FF} 3.93}                       & {\color[HTML]{CB0000} 63.00}  & 81.00                         & 87.00                         & {\color[HTML]{3531FF} 90.00} \\
    RMSO-ConvNeXt\cite{RMSO-ConvNeXt} & 5.59                         & 45.70                         & 66.49                        & 75.74                         & 82.21\\
    Ours                      & {\color[HTML]{CB0000} 2.93}   & {\color[HTML]{3531FF} 62.14} & {\color[HTML]{CB0000} 82.25} & {\color[HTML]{CB0000} 89.19} & {\color[HTML]{CB0000} 93.04} \\ \bottomrule
    \end{tabular}
    }
    
    \label{tab:SEN1-2}
\end{table}

\begin{table}[t]
    \caption{Comparison results of the average $L_{2}$ distance and CMR on the OS:512 dataset. The top two values are marked as red and blue, respectively.}
    \centering
    \resizebox{\linewidth}{!}{
    \begin{tabular}{@{}lcccccc@{}}
    \toprule
                              &                               & \multicolumn{4}{c}{CMR(T) (\%) $\uparrow$}                                                                                                                              \\ \cmidrule(l){3-6} 
    \multirow{-2}{*}{Methods} & \multirow{-2}{*}{$L_{2}$ $\downarrow$} & $T=1$                          & $T=2$                          & $T=3$                          & $T=5$                          \\ \cmidrule(r){1-6}
    Him-net\cite{HimNet}                   & -                             & 27.36                        & 45.75                        & 69.34                        & -                            \\
    DC-InfoNCE\cite{DC-InfoNCE}                & -                             & {\color[HTML]{3531FF} 34.67} & {\color[HTML]{3531FF} 58.96} & {\color[HTML]{3531FF} 78.30} &  92.69 \\
    {\color{blue}OSMnet}\cite{OSMNet}  & {\color[HTML]{CB0000}2.20}                             & 29.25                        & 56.84                        & 77.83                      & {\color[HTML]{3531FF}93.87} \\ 
    F3Net\cite{F3Net}                    &3.18                             & 24.53                            & 51.18                            & 72.41                            & 89.62                            \\
    Ours                      & {\color[HTML]{3531FF}  2.39}   & {\color[HTML]{CB0000} 34.67} & {\color[HTML]{CB0000} 61.32} & {\color[HTML]{CB0000} 80.42} & {\color[HTML]{CB0000} 94.58} \\ \bottomrule
    \end{tabular}
    }
    \label{tab:OS-512}
\end{table}

\begin{figure}[t]
    \centering
    \includegraphics[width=\linewidth]{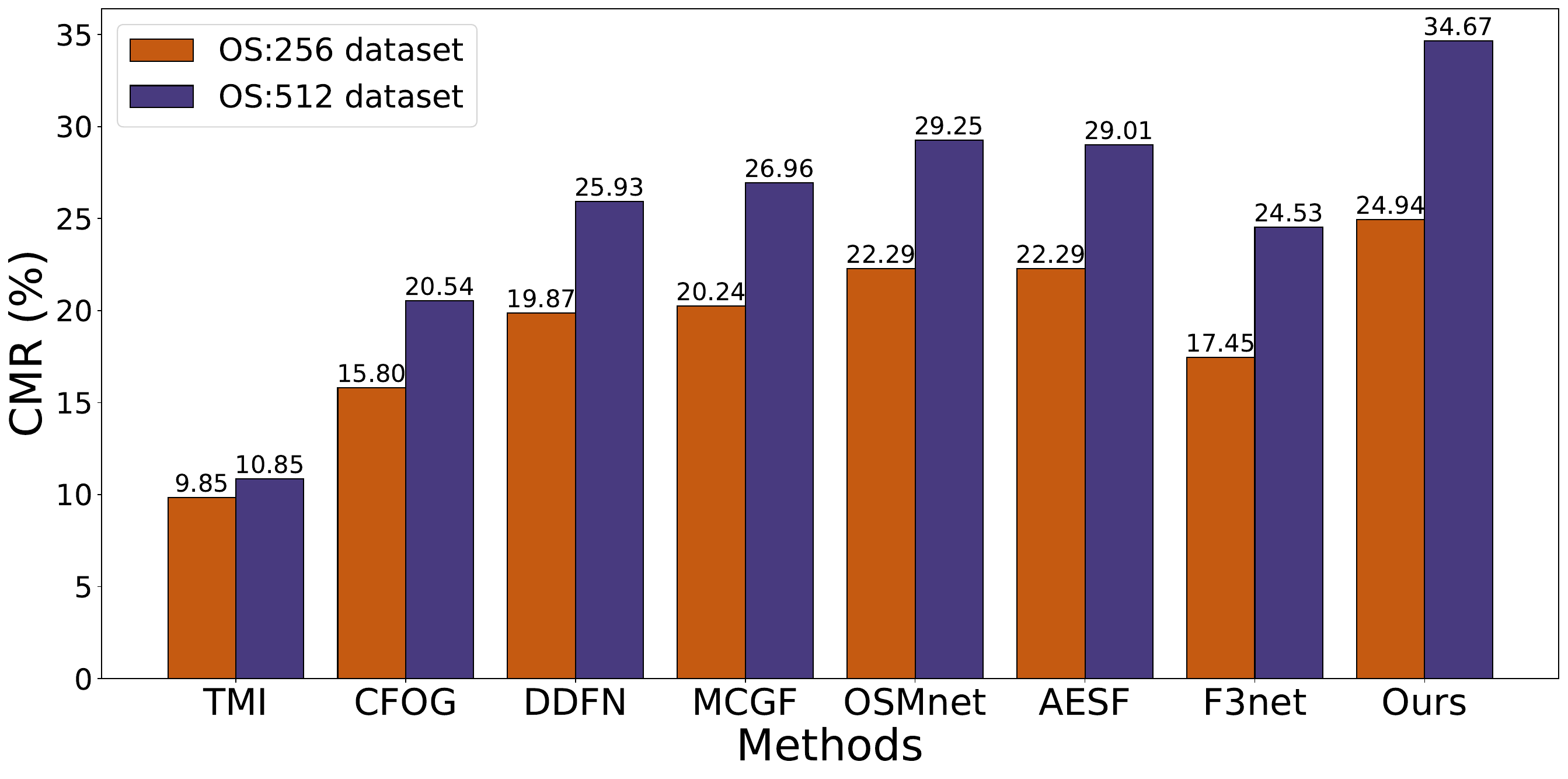}
    \caption{The registration results of various methods at $T=1$ on the OS:256 and OS:512 datasets. Their template image sizes are $192 \times 192$ and $384 \times 384$, respectively.}
    \label{fig:OSdataset_512_256}
\end{figure}

\subsection{Dataset}

This paper tests the effectiveness of the proposed method on two cross-modal image datasets, the SEN1-2 dataset and the OS dataset, which have different image resolutions.

\textbf{SEN1-2 dataset} \cite{SEN1-2} has 282,384 pairs of co-registered optical and SAR image patches acquired from Sentinel-1 and Sentinel-2 satellites. The image resolution is 10 m, and the image size is $256\times 256$. The SEN1-2 dataset contains diverse land cover types (e.g., urban areas, plains, water bodies, deserts) and seasonal variations (spring, summer, autumn, winter). Following the experimental settings in \cite{MARU-Net}, we select 6,450 image pairs from each season and acquire 25,800 for deep network training and testing. The ratio of the number of image pairs for training and testing is $7:3$. Thus, there are 18,060 image pairs for training and 7,740 for testing. We randomly crop a $192\times 192$ image patch from SAR images as the template, where the top-left coordinates of the crop location are the ground truth of matching.

\textbf{OS dataset} \cite{OSdataset} also includes many SAR images and corresponding optical images. The SAR images are acquired using the multi-polarized C-band SAR satellite sensor GF-3 (Gaofen-3), while the optical images are obtained from the Google Earth platform. This dataset covers various cities and plain regions worldwide, such as Beijing and Shanghai in China, Guam and Tucson in the United States, and Agra in India. The OS dataset contains 2,673 non-overlapping image pairs of size $512\times512$ (labeled as OS:512) and 10,692 non-overlapping image pairs of size $256\times256$  (labeled as OS:256). The image resolution is 1 m. For the OS:512 dataset, we use 2,249 image pairs for training and 424 image pairs for testing, and randomly crop $384\times384$ image patches from the SAR images as template images. For the OS:256 dataset, we use 8,996 image pairs for training and 1,696 image pairs for testing, and randomly crop $192\times192$ image patches from the SAR images as template images. 

The main challenges of these two datasets can be summarized as follows:

\textbf{Significant modality differences:} There are significant differences between optical and SAR images caused by the different imaging principles. It is challenging for the model to simultaneously learn shared features invariant to different image modalities and accurately identify registration regions.

\textbf{Fewer local texture features:} The resolution of the SEN1-2 dataset is 10 m, while the resolution of the OS dataset is 1 m. Within a local receptive field of the same size, the OS dataset may contain fewer discriminative textures than the SEN1-2 dataset, significantly increasing the difficulty of achieving precise alignment.

\subsection{Evaluation Metrics}
The $L_{2}$ distance between the predicted matching location and the ground truth can be used to evaluate the image registration accuracy. Additionally, we adopt the Correct Matching Rate (CMR) to test the whole registration performance. CMR represents the successful matching ratio in which the $L_{2}$ distance between the predicted matching location and the ground truth is smaller than a given threshold $T$, denoted as CMR($T$). They can be formulated as follows:

\begin{equation}
\begin{array}{l}  
    CMR(T)=\frac{N_{m}}{N_{t}},\\\\
    L_{2}=\sqrt{(x_{p}-x_{g})^{2}+(y_{p}-y_{g})^{2}}, 
    \end{array}
\end{equation}
where $N_{m}$ represents the number of successful matching image pairs ($L_{2}\le T$), $N_{t}$ is the number of test image pairs, $T$ is the distance threshold, $(x_{p}, y_{p})$ is the predicted matching position, and  $(x_{g}, y_{g})$ is the corresponding ground truth.

\begin{table}[t]
  \centering
  \caption{Ablation experiments on the OS:256 dataset.}
  \resizebox{\linewidth}{!}{
    \begin{tabular}{c|cccccc}
    \toprule
    \multirow{2}[3]{*}{Base} & \multirow{2}[3]{*}{MELF} & \multirow{2}[3]{*}{Mamba}  & \multirow{2}[3]{*}{$L_{2}$ $\downarrow$} & \multicolumn{3}{c}{CMR(T) (\%)  $\uparrow$} \\
\cmidrule{5-7}          &       &       &       & $T=1$   & $T=2$   & $T=3$ \\
    \midrule
    $\checkmark$     &       &       & 4.39  & 19.69  & 37.32  & 56.43  \\
         &$\checkmark$       &      & 3.69  & 23.00    & 44.52 & 65.04 \\
         &      & $\checkmark$      & 3.93  & 20.75 & 42.28 & 62.85 \\
        & $\checkmark$     & $\checkmark$     & 3.39  & 24.94 & 48.11 & 67.92 \\
    \bottomrule
    \end{tabular}%
    }
  \label{tab:key modules}%
\end{table}%

\begin{table}[t]
  \centering
  \caption{Ablation experiments on the MFA module based on the OS:256 dataset.}
  \resizebox{.9\linewidth}{!}{
    \begin{tabular}{lcccc}
    \toprule
    \multicolumn{1}{c}{\multirow{2}[3]{*}{Model}} & \multirow{2}[3]{*}{$L_{2}$ $\downarrow$} & \multicolumn{3}{c}{CMR(T) (\%)  $\uparrow$} \\
\cmidrule{3-5}          &       & $T=1$   & $T=2$   & $T=3$ \\
    \midrule
    
    W/O MSAA & 3.65  & 23.82 & 45.34 & 65.57 \\
    W/O CA & 3.46  & 24.41 & 45.70  & 66.63 \\
    W/O MSAA-CA & 3.49  & 24.88 & 44.75 & 65.15 \\
    Ours & 3.39  & 24.94 & 48.11 & 67.92 \\
    \bottomrule
    \end{tabular}%
    }
  \label{tab:MSAA CA modules ablation}%
\end{table}%

\subsection{Implementation Details}
The proposed ME-SSM is implemented using the PyTorch framework and trained on Nvidia RTX 4090 GPUs. 
We utilize the AdamW optimizer with a batch size of $4$ and an initial learning rate of $5 \times 10^{-4}$. 
The model is trained for $10$ epochs.

\subsection{Comparative experiments}
This section compares the registration results of our ME-SSM with representative handcrafted and deep learning-based methods, including traditional template matching methods, i.e., NCC \cite{Standard_NCC}, MI \cite{MI}, TMI \cite{MI/2} (Truncated MI), CFOG \cite{CFOG}, and  deep feature learning methods, i.e., Siamese CNN \cite{dilated_convolutional}, OSMnet \cite{OSMNet}, MARU-Net \cite{MARU-Net}, Him-Net \cite{HimNet}, DC-InfoNCE \cite{DC-InfoNCE}, F3Net \cite{F3Net}, AESF \cite{AESF}, VU-Net \cite{VUNet} and RMSO-ConvNeXt \cite{RMSO-ConvNeXt}. Note that the results of RMSO-ConvNeXt are obtained by using its released model.

Table \ref{tab:SEN1-2} presents the registration performance of various methods on the SEN1-2 dataset. The proposed ME-SSM achieves the lowest average $L_{2}$ distance and the highest registration accuracy (CMR) across different distance thresholds. Due to the significant differences between optical and SAR images, handcrafted methods have large registration errors. For example, the average $L_{2}$ distance of MI is 11.80, and its CMR(3) is  54.00\%. Deep learning-based methods demonstrate significant advantages over traditional handcrafted methods. These experimental results demonstrate that deep learning methods can better learn shared feature representations across image modalities. 

Our proposed method outperforms all other methods, which acquires an average $L_{2}$  distance of 2.93 and registration accuracy of 89.19\% at $T=3$. F3Net obtains an average $L_{2}$ distance of 4.15 and CMR(3) of 87.24\%. VU-Net achieves the average $L_{2}$ distance of 3.93 and acquires the CMR(3) of 87.00\%. Compared with MARU-Net, VU-Net, and F3Net, our ME-SSM increases CMR(3) by 7.19\%, 2.19\%, and 1.95\%, respectively. Existing deep learning-based methods mainly use the CNN network for feature learning, which has a limited local receptive field and fails to capture global features for image matching and registration. However, our ME-SSM can efficiently capture global contextual information with linear computational complexity through a multi-directional scanning mechanism. Additionally, ME-SSM further enhances the features through multi-level feature aggregation.

Table \ref{tab:OS-512} presents the registration performance of various methods on the OS dataset. DC-InfoNCE achieves a registration accuracy CMR(3) of 78.30\%, while our method achieves CMR(3) of 80.42\% and acquires a performance gain of 2.12\%. Although the images in the OS dataset have limited local textures, our proposed method still acquires better registration performance. ME-SSM can effectively capture both local details and global contextual information, enhancing the richness and discriminability of features for more accurate image registration.

To further demonstrate the performance of our method in precise registration, we present the registration accuracy at $T=1$, CMR(1), based on the SEN1-2 and OS datasets, as shown in Table \ref{tab:SEN1-2} and Fig. \ref{fig:OSdataset_512_256}. It can be seen that our proposed method acquires the best CMR(1) on the SEN1-2 and OS datasets. Specifically, on the SEN1-2 dataset, the best CMR(1) of the traditional handcrafted methods (MI) is 30.00\%, the CMR(1) of the deep feature learning methods (MARU-Net) is 55.00\%, and the accuracy of our method is 62.14\%. ME-SSM acquires performance gains of 32.14\% and 7.14\% compared to traditional and deep learning-based methods, respectively. Although the CMR(1) of VU-Net is slightly higher by 0.86\% than that of our method, its other metrics are significantly lower. On the OS:256 and OS:512 datasets, the best CMR(1) of traditional handcrafted methods (CFOG) are 15.80\% and 20.54\%, respectively. Among deep learning methods, OSMnet demonstrates competitive performance with CMR(1) of 22.29\% and 29.48\%, surpassing or matching AESF (22.29\% and 29.01\%), which explicitly combines deep features and structural features. Our proposed method achieves the best registration results, and its CMR(1) on the OS:256 and OS:512 datasets are 24.94\% and 34.67\%, respectively. Compared to OSMnet, the registration results are improved by 2.65\% and 5.19\%, respectively. 
The superior performance of our proposal is mainly attributed to the following aspects. Firstly, ME-SSM has a strong feature representation ability, which can effectively capture global contextual information and multi-level local features. Secondly, ME-SSM uses multi-expert feature learning to capture salient information across different transformations, greatly enriching the feature representations and boosting image registration performance on remote sensing images with limited textures.

\begin{table*}[t]
  \centering
  \caption{Comparison experiments on the OS:256 dataset. ``$\dag$" represents the deep model that uses the multi-expert learning framework. The best results are marked in red, and the performance gains are represented in green.}
  \resizebox{0.9\linewidth}{!}{
    \begin{tabular}{lcccccccc}
    \toprule
    \multirow{2}[4]{*}{Methods} & \multirow{2}[4]{*}{$L_{2}$ $\downarrow$} & \multicolumn{5}{c}{CMR(T) (\%)  $\uparrow$}           & \multirow{2}[4]{*}{Params (M)} & \multirow{2}[4]{*}{FLOPs (G)} \\
\cmidrule{3-7}          &       & $T=1$   & $T=2$   & $T=3$   & $T=4$   & $T=5$   &       &  \\
    \midrule
    UNet\cite{U-net}  & 4.39  & 19.69  & 37.32  & 56.43  & 69.93  & 79.60  & 2.44  & 48.82  \\
    UNet$\dag$ & \cellcolor{green!10}3.69 & \cellcolor{green!10}23.00 & \cellcolor{green!10}44.52 & \cellcolor{green!10}65.04 & \cellcolor{green!10}77.30 & \cellcolor{green!10}84.91 & 2.55  & 216.30  \\
    \midrule
    TransUNet\cite{Transunet} & 4.04  & 19.93  & 42.16  & 62.50  & 74.59  & 82.31  & 210.57  & 120.44  \\
    TransUNet$\dag$ & \cellcolor{green!10}3.39 & \cellcolor{green!10}24.35 & \cellcolor{green!10}45.93 & \cellcolor{green!10}65.63 & \cellcolor{green!10}77.18 & \cellcolor{green!10}84.67 & 210.64  & 489.35  \\
    \midrule
    U-transformer\cite{U-net_transformer} & 8.16  & 18.99  & 36.44  & 56.72  & 66.63  & 74.00  & 23.36  & 319.01  \\
    U-transformer$\dag$ & \cellcolor{green!10}5.92 & \cellcolor{green!10}18.81 & \cellcolor{green!10}38.56 & \cellcolor{green!10}59.20 & \cellcolor{green!10}70.58 & \cellcolor{green!10}78.60 & 23.43  & 1283.64  \\
    \midrule
    SwinTransUnet\cite{Swintransformer} & 4.87  & 19.69  & 39.27  & 58.02  & 69.99  & 78.48  & 27.08  & 31.07  \\
    SwinTransUnet$\dag$ & \cellcolor{green!10}3.77  & \cellcolor{green!10}23.23  & \cellcolor{green!10}43.69  & \cellcolor{green!10}63.38  & \cellcolor{green!10}76.89  & \cellcolor{green!10}84.38  & 27.15  & 131.88  \\
    \midrule
    VmambaUnet\cite{VMamba} & 4.81  & 20.22  & 39.98  & 57.90  & 70.22  & 79.25  & 21.08  & 27.13  \\
    VmambaUnet$\dag$ & \cellcolor{green!10}3.80  & \cellcolor{green!10}22.64  & \cellcolor{green!10}43.99  & \cellcolor{green!10}63.15  & \cellcolor{green!10}75.29  & \cellcolor{green!10}83.79  & 22.94  & 121.17  \\
    \midrule
    F3Net\cite{F3Net} & 5.92  & 17.75  & 34.55  & 55.07  & 67.33  & 75.65  & 0.28  & 35.73  \\
    F3Net$\dag$ & \cellcolor{green!10}5.37  & \cellcolor{green!10}18.04  & \cellcolor{green!10}35.85  & \cellcolor{green!10}57.31  & \cellcolor{green!10}69.34  & \cellcolor{green!10}77.24  & 0.55  & 223.29  \\
    \midrule
    Ours  & {\color[HTML]{CB0000} 3.39} & {\color[HTML]{CB0000} 24.94} & {\color[HTML]{CB0000} 48.11} & {\color[HTML]{CB0000} 67.92} & {\color[HTML]{CB0000} 78.60} & {\color[HTML]{CB0000} 85.50} & 22.14  & 170.24  \\
    \bottomrule
    \end{tabular}%
    }
  \label{tab:MELF}%
\end{table*}%

\begin{figure}[t] 
    \centering
    \includegraphics[width=\linewidth]{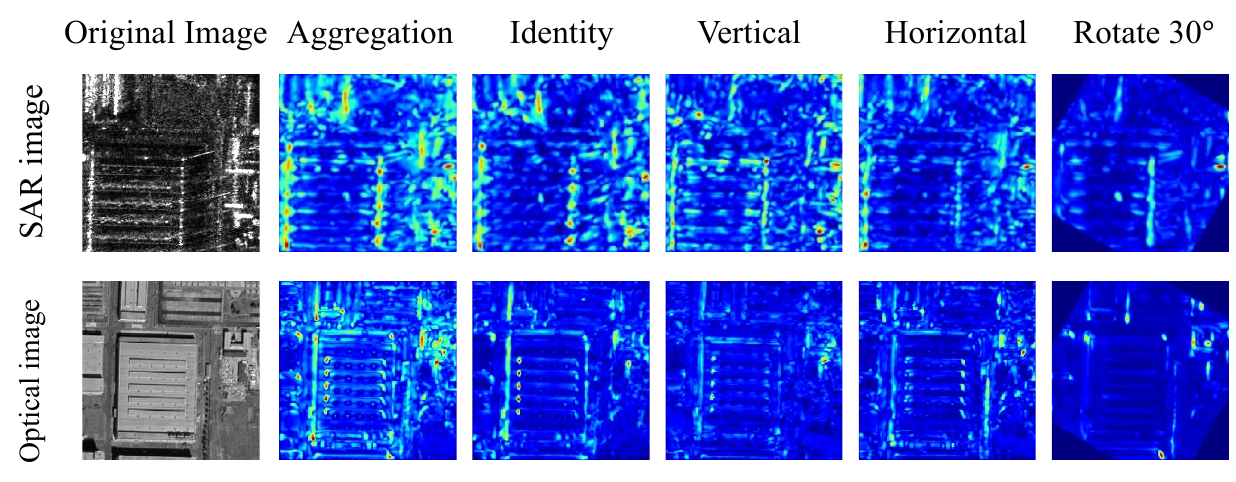} 
    \caption{Feature maps extracted by different experts under high-intensity geometric transformations, where ``Aggregation'' represents the feature aggregated by the soft router. The image transformations are Identity(identity transformation), Horizontal(horizontal flip), Vertical(vertical flip), and Rotation 30°(high-intensity rotation transformation).}
    \label{fig:each_expert}
\end{figure}

\subsection{Influence of the MELF and Mamba}
This section explores the effectiveness of the crucial modules in our proposed ME-SSM based on the OS:256 dataset, such as the MELF and Mamba framework. Experimental results are presented in Table \ref{tab:key modules}. ``Base" refers to the U-shaped network based on the CNN framework. ``MELF" denotes the deep model that uses our proposed multi-expert learning framework.

Experimental results in Table \ref{tab:key modules} demonstrate the effectiveness of our proposed MELF and Mamba framework. When applying the MELF, the average $L_{2}$ decreased from 4.39 to 3.69, and the CMR(3) increased from 56.43\% to 65.04\%. Meanwhile, our Mamba framework acquires registration performance gains compared to the Base framework. When using the Mamba for feature extraction, the average $L_{2}$ decreases to 3.93, and the CMR(3) increases to 62.85\%. Our proposal with the MELF and Mamba achieves the best registration results, and its average $L_{2}$ is 3.39, and CMR(3) is 67.92\%.

\subsection{Influence of the MFA}
We also conduct ablation studies on the MFA (MSAA and CA modules) in ME-SSM, and the results are presented in Table \ref{tab:MSAA CA modules ablation}. ``W/O MSAA" denotes the deep network that does not contain the MSAA module, and ``W/O MSAA-CA" denotes the deep network does not contain the MSAA and CA modules. 

The experimental results indicate that the MSAA and CA modules can enhance registration performance. Specifically, compared to our model with MSAA and CA, the ``W/O MSAA", ``W/O CA", and ``W/O MSAA-CA"  increase the average $L_{2}$ by 0.26, 0.07, and 0.10, respectively, and decrease CMR(3) by 2.35\%, 1.29\%, and 2.77\%, respectively. 

The MSAA module adaptively extracts multi-scale features at each level in the encoder, enhancing the local texture information. The CA module adaptively reallocates channel features within the high-dimensional latent space. The MFA combines the MSAA and CA modules and encourages ME-SSM to effectively extract global contextual information while preserving the importance of local information, facilitating image registration performance.

\begin{figure}[t] 
    \centering
    \includegraphics[width=\linewidth]{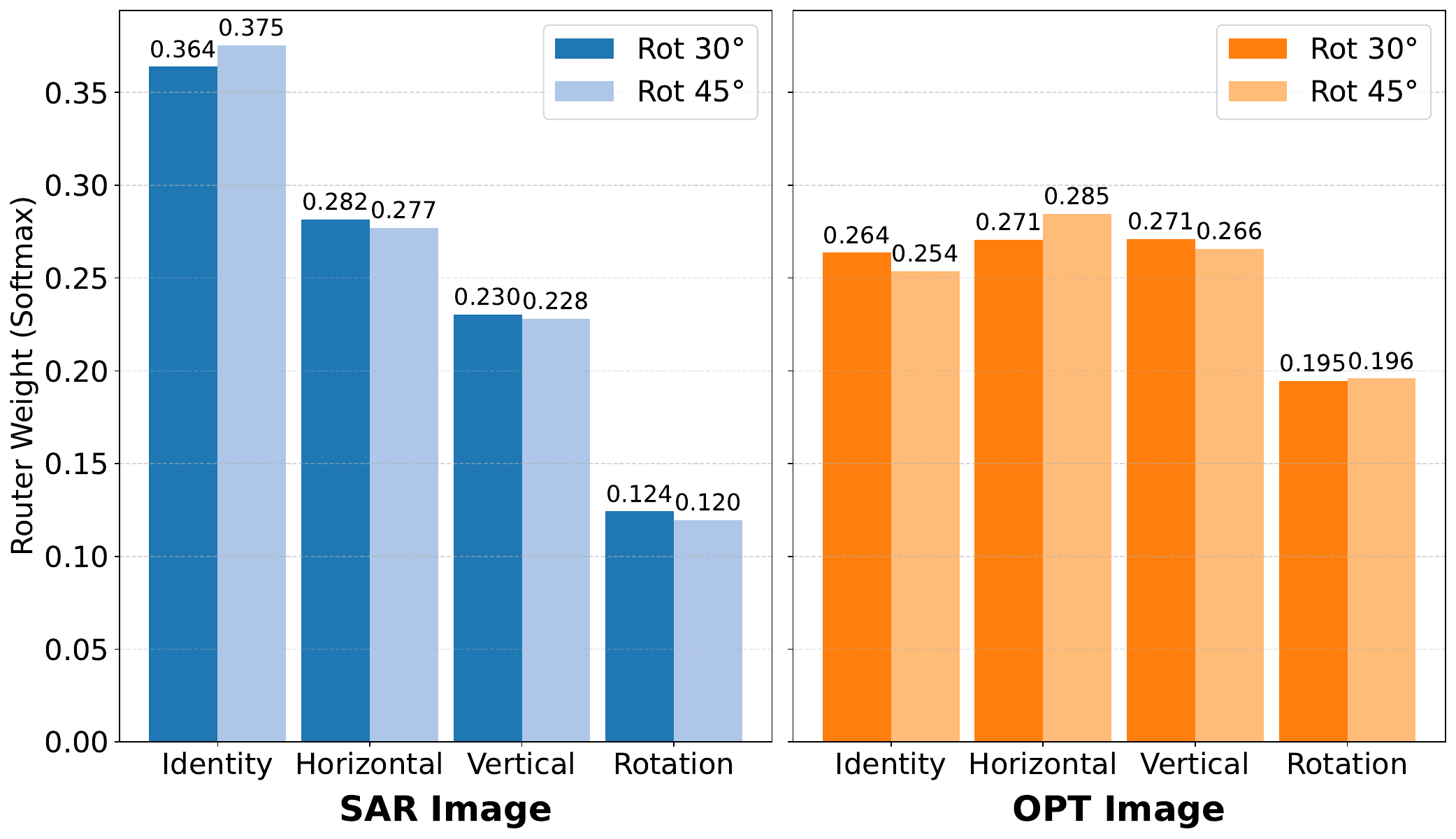} 
    \caption{Router weights of experts under different image transformations. The four image transformations are Identity (Identity transformation), Horizontal (horizontal flip), Vertical (vertical flip), and Rotation (rotation transformation).}
    \label{fig:rot}
\end{figure}

\begin{table}[t]
  \caption{Ablation experiments on the number of experts and fusion method based on the OS:256 dataset. }
    \centering
    \resizebox{.9\linewidth}{!}{
    \begin{tabular}{lcccc}
    \toprule
    \multirow{2}[3]{*}{Method} & \multirow{2}[3]{*}{$L_{2}$ $\downarrow$} & \multicolumn{3}{c}{CMR(T) (\%)  $\uparrow$} \\
\cmidrule{3-5}          &       & $T=1$   & $T=2$   & $T=3$ \\
    \midrule
    MELF-0 & 3.93  & 20.75 & 42.28 & 62.85 \\
    MELF-2 & 3.70  & 23.00 & 43.40 & 65.04 \\
    MELF-3 & 3.53  & 24.65 & 45.17 & 66.16 \\
    MELF-4 & {\color[HTML]{CB0000} 3.39}  & {\color[HTML]{CB0000} 24.94} & {\color[HTML]{CB0000} 48.11} & {\color[HTML]{CB0000} 67.92} \\
    MELF-5 & 3.41  & 24.29 & 46.46 & 67.04 \\
    \midrule
    MELF-4-rot30 & 3.44  & 24.76 & 46.22 & 65.92 \\
    MELF-4-rot45 & 3.68  & 23.34 & 43.81 & 65.04 \\
    \midrule
    MELF-S-2(4) & 3.69  & 22.41 & 44.22 & 64.86 \\
    MELF-4-A & 3.50  & 24.06 & 46.52 & 66.63 \\
    \bottomrule
    \end{tabular}%
    }
  \label{tab:ablation of experts number}%
\end{table}%

\begin{figure*}[t] 
    \centering
    \includegraphics[width=\textwidth]{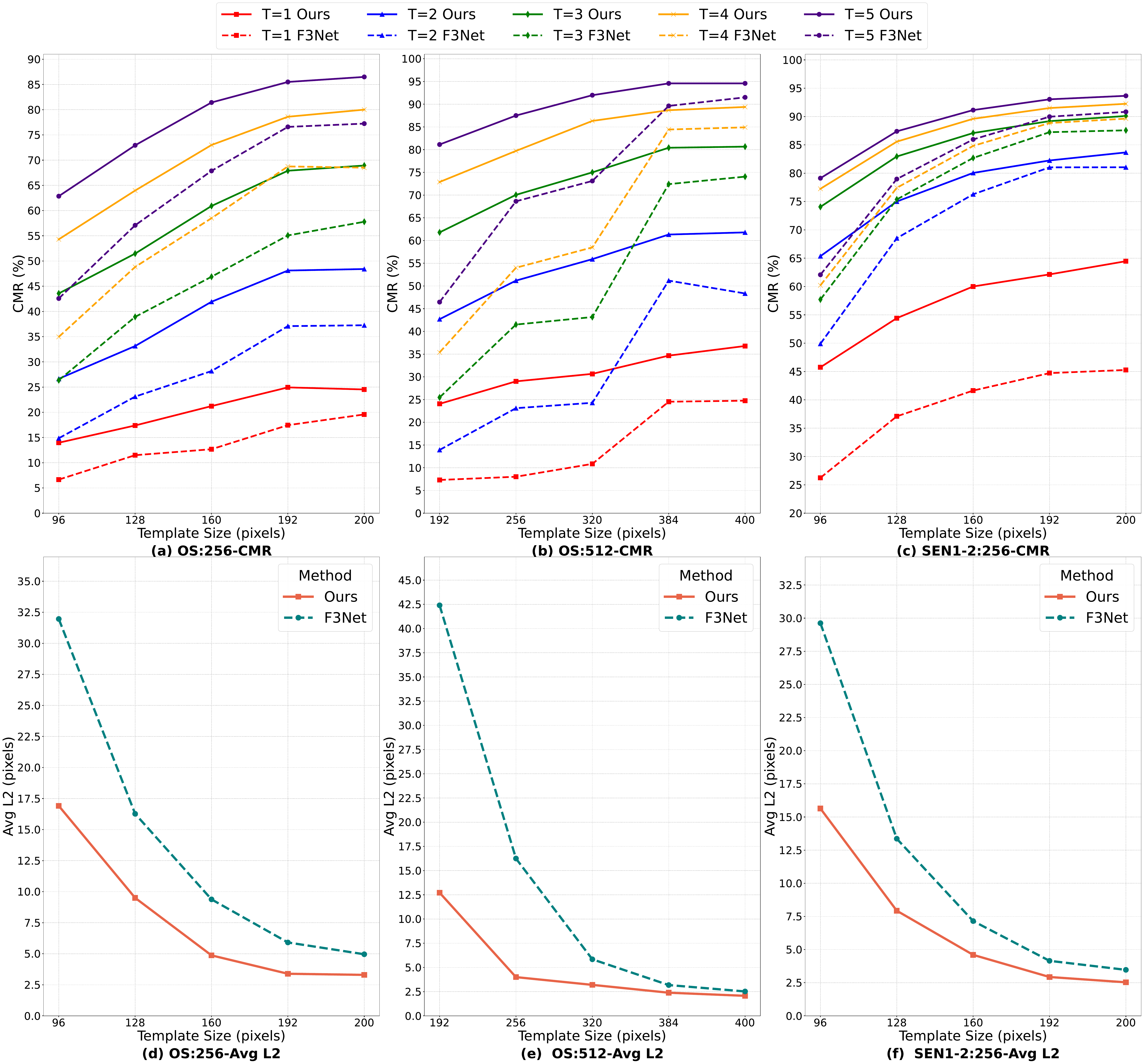} 
    \caption{Registration performance of our ME-SSM and F3Net on the OS dataset and SEN1-2 dataset under different template sizes and threshold values.}
    \label{fig:parameters_test}
\end{figure*}

\subsection{The effectiveness of MELF on other frameworks}

This section validates the effectiveness of our proposed multi-expert learning on other frameworks, such as UNet\cite{U-net}, the Transformer-based methods of TransUNet \cite{Transunet}, Swin Transformer \cite{Swintransformer}, and the SSM-based Vmamba \cite{VMamba}.

Firstly, MELF can effectively improve image registration performance based on various registration frameworks. As shown in Table \ref{tab:MELF}, when the MELF is applied to UNet, the average $L_{2}$ decreases by 0.70 (from 4.39 to 3.69), and the CMR(3) improves by 8.61\% (from 56.43\% to 65.04\%). Similarly, when the MELF is applied to TransUNet, the average $L_{2}$ decreases by 0.65 (from 4.04 to 3.39), and the CMR(3) increases by 3.13\% (from 62.50\% to 65.63\%). Furthermore, MELF demonstrates strong generalizability on the latest architectures. For instance, when we integrate MELF into VmambaUnet and SwinTransUnet, their CMR(3) improves by 5.25\%  (from 57.90\% to 63.15\%) and 5.36\% (from 58.02\% to 63.38\%), respectively. To further validate the effectiveness of MELF, we conducted experiments on F3Net, a representative frequency domain-based method in optical-SAR image registration. Our MELF increases the  CMR(3)  of F3Net by  2.24\% and decreases the $L_2$ distance by 0.55.

These results demonstrate that our MELF is effective and universal, seamlessly integrating into various frameworks and yielding significant performance improvements.

Secondly, Transformer-based frameworks acquire better cross-modal image registration results. For example, UNet achieves an average $L_{2}$ of 4.39 and a CMR(3) of 56.43\%. However, TransUNet achieves an average $L_{2}$ of 4.04, with a CMR(3) of 62.50\%. Compared to UNet, TransUNet benefits from its self-attention mechanism and global receptive field, reducing the average $L_{2}$ by 0.35 and improving the CMR(3) by 6.07\%. However, Transformer-based frameworks significantly increase computational costs. For example, the number of parameters of UNet is 2.44M, while the number of parameters of TransUNet is 210.57M.

Additionally, our ME-SSM has advantages over other methods in terms of registration performance and computation complexity. Our ME-SSM achieves an average $L_{2}$ of 3.39 and a CMR(3) of 67.92\%. Compared to UNet+MELF, our method reduces the average $L_{2}$ by 0.3 and improves the CMR(3) by 2.88\%. Compared to TransUNet+MELF, our method improves the CMR(3) by 2.29\%.  Additionally, compared to the strong baselines such as SwinTransUnet+MELF and VmambaUnet+MELF, our method presents superior performance gains by 4.54\% and 4.77\% in CMR(3), respectively. Notably, ME-SSM achieves optimal performance with only 22.14M network parameters and 170.24G FLOPs. The reason is that ME-SSM can efficiently capture global contextual information with linear complexity, achieving good registration performance without expensive computational burdens. Furthermore, the MELF enhances the richness and discriminability of features, contributing to significant performance gains in images with limited textures.

\subsection{Influence of the number of experts and fusion method in MELF}
This paper adopts the MELF to extract rich features from images. This section investigates the impact of the number of feature experts and the fusion method in the MELF on registration performance. As shown in Table \ref{tab:ablation of experts number}, ``MELF-0" denotes our ME-SSM without the MELF. ``MELF-2" and ``MELF-5" represent ME-SSM fuses features from the 2 and 5 experts, respectively. As the number of experts increases, the ME-SSM achieves lower average $L_{2}$ and higher CMR. ``MELF-4" acquires the best registration performance. Compared to ``MELF-0", the average $L_{2}$ of ``MELF-4" is reduced by 0.54, and CMR(3) of ``MELF-4" is increased by 5.07\%. 

This part also simulates a sparse gated fusion of expert features in MELF. For example, ``MELF-S-2(4)" indicates that MELF only selects two expert features from four experts for fusion. In comparison to ``MELF-4", the results of ``MELF-S-2(4)" showed an increase in average $L_{2}$ by 0.3 and a decrease in CMR(3) by 3.06\%. Therefore, we focus on integrating knowledge from all experts to enrich feature representation and enhance feature discriminability. Additionally, we compare our soft router dynamic fusion method and the hard gating expert feature fusion mechanism. ``MELF-4-A" treats all expert features equally. It can be seen that our learnable soft router fusion mechanism performs better than ``MELF-4-A". Specifically,  the average $L_{2}$ decreases by 0.11, and the CMR(3) increases by 1.29\%. This improvement is attributed to each expert focusing on different patterns of features. The learnable soft router selectively filters out task-irrelevant features and fuses important salient features for image registration.

Furthermore, we investigated the impact of large transformations on registration performance. Following the same transformation setting in the multi-expert learning framework, we also adopt four experts and replace the ``central rotation'' (without interpolation) with significant rotations, i.e., rotations of $30^\circ$ and $45^\circ$, which were labeled as ``MELF-4-rot30'' and ``MELF-4-rot45'', respectively. Experimental results on the OS:256 dataset, as presented in Table \ref{tab:ablation of experts number}, demonstrate that the large geometric transformation brings a performance degradation. Specifically, when we use rotations $30^\circ$ and $45^\circ$ for input image transformation, the CMR(3) decreases by 2.00\% and 2.88\%, respectively, and the average $L_{2}$ distance increases by 0.05 and 0.29, respectively. 
This demonstrates that while geometric augmentation is generally beneficial, strong transformations may induce information loss, impairing fine-grained image registration performance. For instance, as illustrated in Figure \ref{fig:each_expert}, the expert captures fewer features from the transformed image with rotation $30^\circ$ than from other transformed images. 
Additionally, we found that our soft router effectively performs adaptive expert feature selection by examining the learned soft routing weights. As shown in Figure \ref{fig:rot}, the router assigned significantly lower weights to the experts handling the large transformation views (rotation $30^\circ$ and $45^\circ$) than identity transformation, horizontal flip, and vertical flip. Thus, we adopt the slight transformation in our MELF framework.

The negative impacts of strong transformations may be attributed to two primary factors: information loss and accumulation of feature errors. Firstly, significant rotations usually adopt bilinear or bicubic interpolation to map pixels to the new grid. The image interpolation operator can be viewed as a low-pass smoothing filter, which will reduce high-frequency details, resulting in information loss and decreasing cross-modal image registration performance. Additionally, we perform image transformation and inverse transformation before and after feature extraction, respectively. The large image transformations will bring cumulative errors on features caused by the interpolation operator in the image transformation and inverse transformation, thereby influencing the feature aggregation and registration accuracy.

\subsection{Influence of the template image size}

This section primarily investigates the impact of template image size on the performance of image registration based on the OS and the SEN1-2 datasets. For the OS:256 and SEN1-2 datasets, the reference image size is $256 \times 256$, the template image sizes are $200 \times 200$, $192 \times 192$, $160 \times 160$, $128 \times 128$, and $96 \times 96$, respectively. For the OS:512 dataset, where the reference images are $512 \times 512$, template image sizes are $400 \times 400$, $384 \times 384$, $320 \times 320$, $256 \times 256$, and $192 \times 192$, respectively. As the template image size decreases, the potential matching localization space increases.

Figure \ref{fig:parameters_test} presents the registration results of our proposal and F3Net. As the size of the template image increases, the corresponding CMR significantly improves, and the average $L_{2}$ gradually decreases. For example, as shown in Fig. \ref{fig:parameters_test} (a) and (d), when the template image size increases from 96 pixels to 128 pixels, CMR(3) improves by 7.90\%, and average $L_{2}$ decreases by 43.80\%. When the template image size increases from 128 to 160 pixels, CMR(3) increases by 9.44\%, and the average $L_{2}$ decreases by 48.70\%. When the template image size increases from 160 to 200 pixels, CMR(3) improves by 8.02\%, while the average $L_{2}$ only decreases by 32.20\%. The main reason is that when the template image size is too small, the image lacks sufficient discriminative information, resulting in poor registration accuracy. Meanwhile, when the reference image size is fixed, the template image with a small size tends to have a wider spatial range for matching, increasing the difficulty of image registration. As the template image size increases, the computational cost also increases.

Additionally, our proposed method significantly outperforms F3Net in registration accuracy under different template image sizes. Moreover, our method is more robust than F3Net to template size changes. For example, as shown in Fig. \ref{fig:parameters_test} (b), when the template size decreases from 384 to 320, the CMR(1) of our method decreases by 4.01\%, while F3Net decreases by 13.68\%; the CMR(2) of our method decreases by 5.42\%, while F3Net decreases by 26.89\%. As shown in Fig. \ref{fig:parameters_test} (e), when the template size decreases from 256 pixels to 192 pixels, the average $L_{2}$ of our method increases by 8.72, while F3Net increases by 26.15. These results further demonstrate the robustness of our method, which can stably adapt to templates of different sizes while maintaining good registration accuracy. Furthermore, our method can extract discriminative features from small template images with sparse textures, thereby improving registration performance.

\begin{figure*}[ht] 
    \centering
    \includegraphics[width=\textwidth]{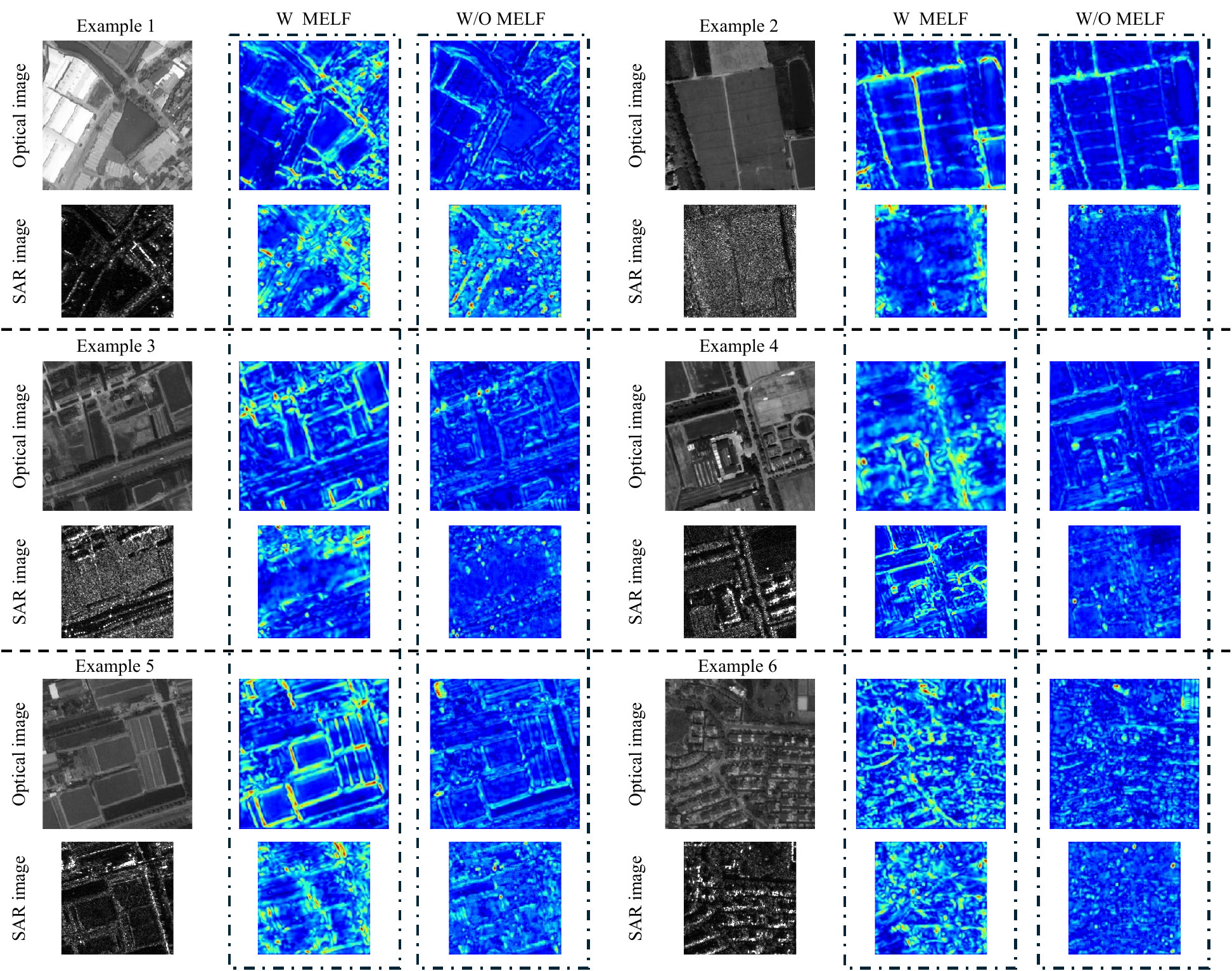} 
    \caption{Feature visualization of optical and SAR images when the deep model with (W) and without (W/O) MELF.}
    \label{fig:ablation_MELF}
\end{figure*}

\begin{figure*}[ht]  
    \centering
    \begin{minipage}{0.16\linewidth}
        \centering
        \includegraphics[width=\linewidth]{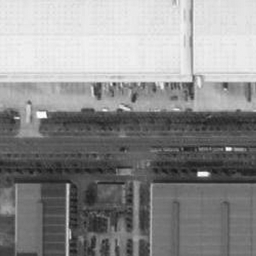}
    \end{minipage}
    \hfill
    \begin{minipage}{0.16\linewidth}
        \centering
        \includegraphics[width=0.75\linewidth]{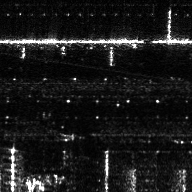}
    \end{minipage}
    \hfill
    \begin{minipage}{0.16\linewidth}
        \centering
        \includegraphics[width=\linewidth]{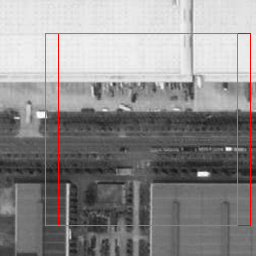}
    \end{minipage}
    \hfill
    \begin{minipage}{0.16\linewidth}
        \centering
        \includegraphics[width=\linewidth]{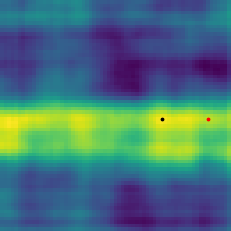}
    \end{minipage}
    \hfill
    \begin{minipage}{0.16\linewidth}
        \centering
        \includegraphics[width=\linewidth]{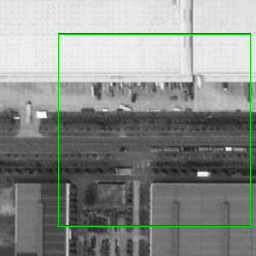}
    \end{minipage}
    \hfill
    \begin{minipage}{0.16\linewidth}
        \centering
        \includegraphics[width=\linewidth]{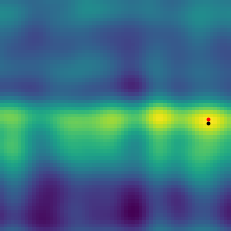}
    \end{minipage}
    
    \vspace{5pt}  

    \begin{minipage}{0.16\linewidth}
        \centering
        \includegraphics[width=\linewidth]{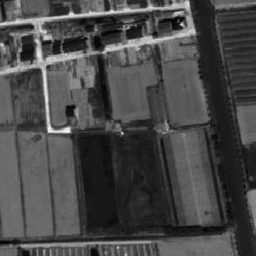}
    \end{minipage}
    \hfill
    \begin{minipage}{0.16\linewidth}
        \centering
        \includegraphics[width=0.75\linewidth]{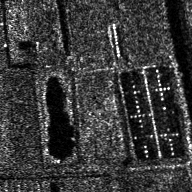}
    \end{minipage}
    \hfill
    \begin{minipage}{0.16\linewidth}
        \centering
        \includegraphics[width=\linewidth]{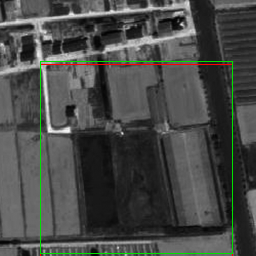}
    \end{minipage}
    \hfill
    \begin{minipage}{0.16\linewidth}
        \centering
        \includegraphics[width=\linewidth]{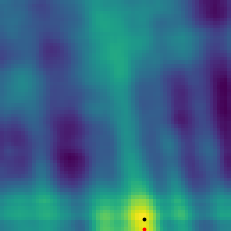}
    \end{minipage}
    \hfill
    \begin{minipage}{0.16\linewidth}
        \centering
        \includegraphics[width=\linewidth]{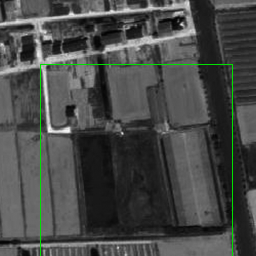}
    \end{minipage}
    \hfill
    \begin{minipage}{0.16\linewidth}
        \centering
        \includegraphics[width=\linewidth]{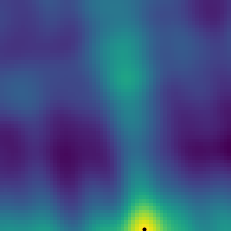}
    \end{minipage}

    \vspace{5pt}  

    \begin{minipage}{0.16\linewidth}
        \centering
        \includegraphics[width=\linewidth]{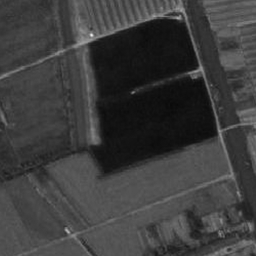}
    \end{minipage}
    \hfill
    \begin{minipage}{0.16\linewidth}
        \centering        \includegraphics[width=0.75\linewidth]{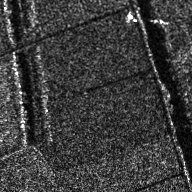}
    \end{minipage}
    \hfill
    \begin{minipage}{0.16\linewidth}
        \centering
        \includegraphics[width=\linewidth]{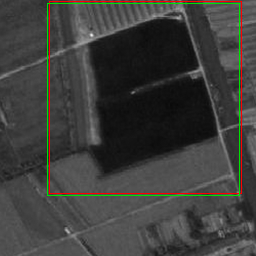}
    \end{minipage}
    \hfill
    \begin{minipage}{0.16\linewidth}
        \centering
        \includegraphics[width=\linewidth]{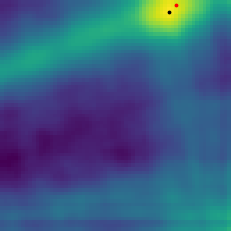}
    \end{minipage}
    \hfill
    \begin{minipage}{0.16\linewidth}
        \centering
        \includegraphics[width=\linewidth]{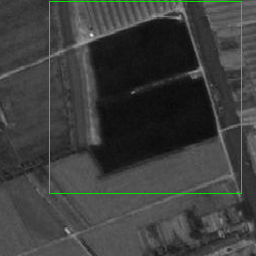}
    \end{minipage}
    \hfill
    \begin{minipage}{0.16\linewidth}
        \centering
        \includegraphics[width=\linewidth]{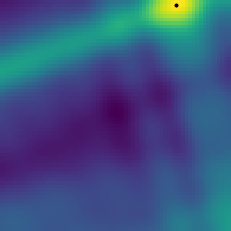}
    \end{minipage}

    \vspace{5pt}

    \begin{minipage}{0.16\linewidth}
        \centering
        \includegraphics[width=\linewidth]{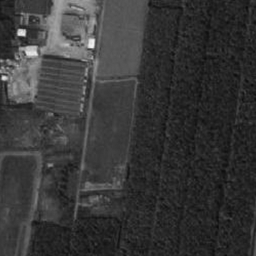}
    \end{minipage}
    \hfill
    \begin{minipage}{0.16\linewidth}
        \centering
        \includegraphics[width=0.75\linewidth]{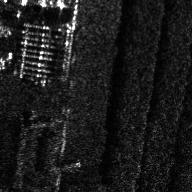}
    \end{minipage}
    \hfill
    \begin{minipage}{0.16\linewidth}
        \centering
        \includegraphics[width=\linewidth]{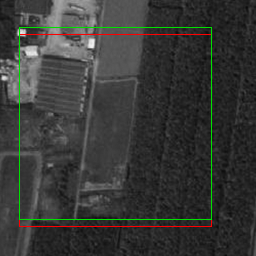}
    \end{minipage}
    \hfill
    \begin{minipage}{0.16\linewidth}
        \centering
        \includegraphics[width=\linewidth]{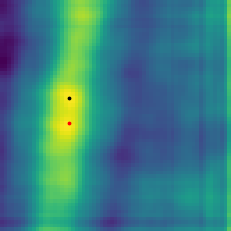}
    \end{minipage}
    \hfill
    \begin{minipage}{0.16\linewidth}
        \centering
        \includegraphics[width=\linewidth]{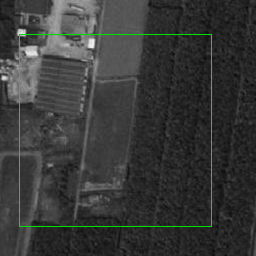}
    \end{minipage}
    \hfill
    \begin{minipage}{0.16\linewidth}
        \centering
        \includegraphics[width=\linewidth]{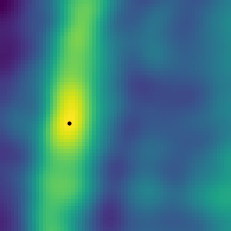}
    \end{minipage}    

    \vspace{5pt}  

    \begin{minipage}{0.16\linewidth}
        \centering
        \includegraphics[width=\linewidth]{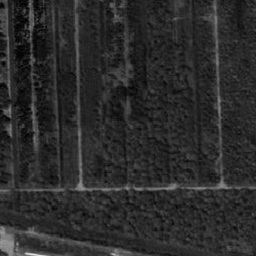}
    \end{minipage}
    \hfill
    \begin{minipage}{0.16\linewidth}
        \centering
        \includegraphics[width=0.75\linewidth]{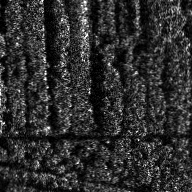}
    \end{minipage}
    \hfill
    \begin{minipage}{0.16\linewidth}
        \centering
        \includegraphics[width=\linewidth]{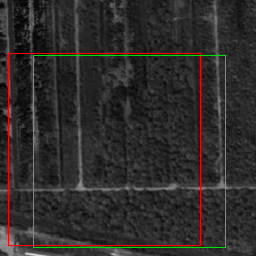}
    \end{minipage}
    \hfill
    \begin{minipage}{0.16\linewidth}
        \centering
        \includegraphics[width=\linewidth]{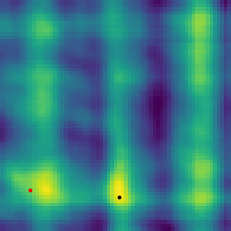}
    \end{minipage}
    \hfill
    \begin{minipage}{0.16\linewidth}
        \centering
        \includegraphics[width=\linewidth]{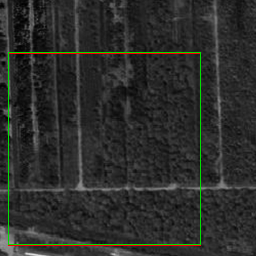}
    \end{minipage}
    \hfill
    \begin{minipage}{0.16\linewidth}
        \centering
        \includegraphics[width=\linewidth]{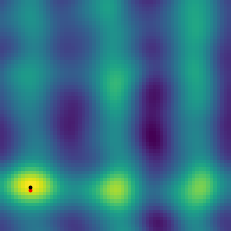}
    \end{minipage}

    \vspace{5pt}  

    \begin{minipage}{0.16\linewidth}
        \centering
        \includegraphics[width=\linewidth]{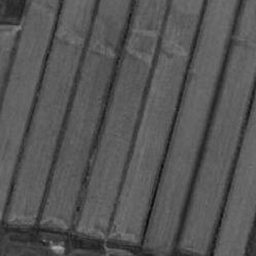}
    \end{minipage}
    \hfill
    \begin{minipage}{0.16\linewidth}
        \centering
        \includegraphics[width=0.75\linewidth]{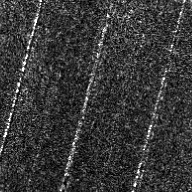}
    \end{minipage}
    \hfill
    \begin{minipage}{0.16\linewidth}
        \centering
        \includegraphics[width=\linewidth]{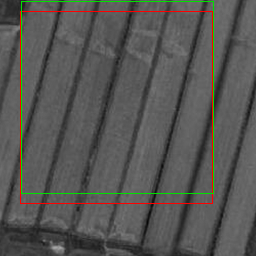}
    \end{minipage}
    \hfill
    \begin{minipage}{0.16\linewidth}
        \centering
        \includegraphics[width=\linewidth]{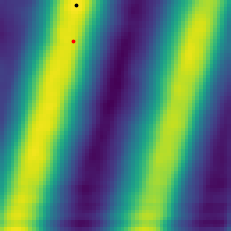}
    \end{minipage}
    \hfill
    \begin{minipage}{0.16\linewidth}
        \centering
        \includegraphics[width=\linewidth]{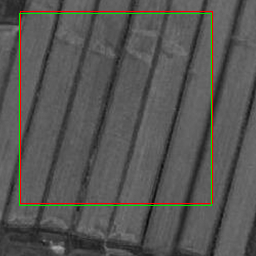}
    \end{minipage}
    \hfill
    \begin{minipage}{0.16\linewidth}
        \centering
        \includegraphics[width=\linewidth]{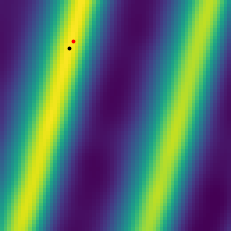}
    \end{minipage}
    \caption{The visual results of F3Net and our ME-SSM on the OS:256 dataset. The first and second columns represent the optical image (reference image) and the corresponding SAR image (template image), respectively. The third and fourth columns are the registration results and similarity maps of F3Net, respectively. The fifth and sixth columns represent the registration results and similarity maps of ME-SSM, respectively.}
    \label{fig:visulization_registration_results}
\end{figure*}

\subsection{Computational efficiency}
This section displays the computational efficiency of our method and other image registration approaches based on the SEN 1-2 dataset, as shown in Table \ref{tab:efficiency}. 

In terms of computational speed, our proposed method achieves the best performance and lower time cost. The average registration time on a pair of images of our method is 123 ms and achieves a CMR(3) of 89.20\%. Specifically, NCC \cite{NCC} is the fastest (57 ms) as it directly computes pixel-level similarity without feature extraction. However, due to the significant radiometric differences in cross-modal images, NCC yields the worst registration accuracy (CMR(3) of 5.00\%). Among deep learning-based methods, Siamese CNN \cite{dilated_convolutional} consumes considerable time (629 ms) due to its heavy feature extraction process. FFT U-Net \cite{FFTUnet} and MARU-Net \cite{MARU-Net} utilize the U-Net framework with frequency domain computations, resulting in inference speeds of 437 ms and 396 ms, respectively. It is worth noting that F3Net \cite{F3Net} achieves remarkable efficiency (71 ms) and the second-best accuracy (87.24\%) by employing a lightweight feature extraction backbone. Although our method (123 ms) is slightly slower than F3Net and NCC, it achieves the highest registration accuracy with a CMR(3) of 89.20\%. This demonstrates that our ME-SSM offers the optimal trade-off between registration accuracy and computational efficiency. The high computational efficiency and performance of our ME-SSM mainly benefit from two aspects. Firstly, it takes advantage of the parallel optimization capabilities of the Mamba framework. Second, we adopt the GPU to accelerate similarity calculations by using the template image as convolution kernels and convolution with the reference image. Moreover, our approach effectively captures global and local features, preserving high accuracy while achieving rapid inference speeds.

\begin{table}[t]
  \caption{The average registration time of a pair of images and CMR(3) of existing methods on the SEN1-2 dataset. The top two values are marked as red and blue.}
  \centering
  \resizebox{\linewidth}{!}{
    \begin{tabular}{lcc}
    \toprule
    Model & Average time (ms) $\downarrow$ & CMR($3$) (\%) $\uparrow$ \\
    \midrule
    NCC\cite{Standard_NCC}    & {\color[HTML]{CB0000} 57}  & 5.00 \\
    MI\cite{MI}    & 5450  & 54.00 \\
    DDFN\cite{DDFN}  & 239   & 50.00 \\
    Siamese CNN\cite{dilated_convolutional} & 629   & 72.00 \\
    FFT U-Net\cite{FFTUnet} & 437   & 74.00 \\
    MARU-Net\cite{MARU-Net} & 396   & 82.00 \\
    F3Net\cite{F3Net} & {\color[HTML]{3531FF} 71}   & {\color[HTML]{3531FF} 87.24} \\
     Ours & 123 & {\color[HTML]{CB0000} 89.20} \\
    \bottomrule
    \end{tabular}%
    }
  \label{tab:efficiency}%
\end{table}%

\begin{figure}[t] 
    \centering
    \includegraphics[width=\linewidth]{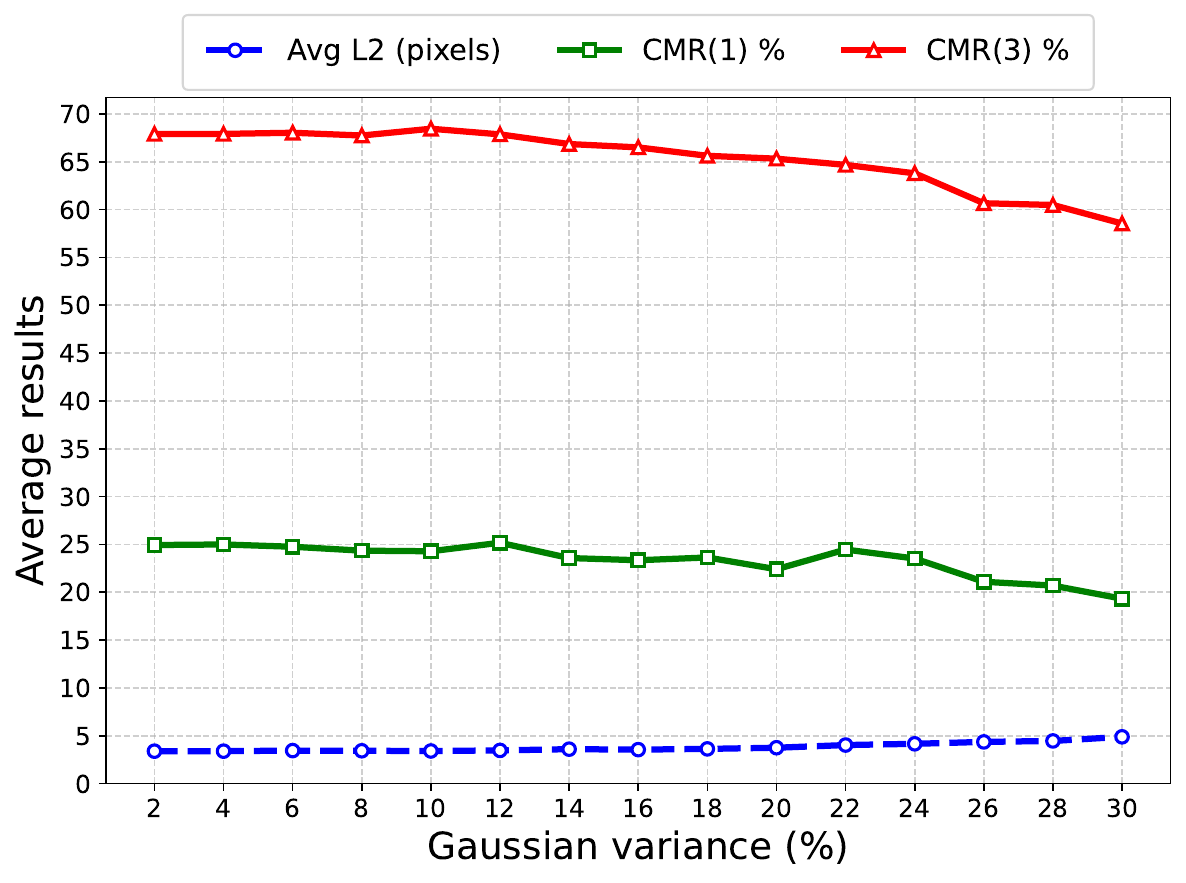} 
    \caption{The performance of ME-SSM under different levels of Gaussian noise.}
    \label{fig:robustness}
\end{figure}

\subsection{Visualization}
\subsubsection{Visualization of features}
This part presents the features of the ME-SSM with (W) and without (W/O) the MELF in Fig. \ref{fig:ablation_MELF}. It can be observed that when the deep model uses the MELF, the features of the optical and SAR images have more distinctive information and sharper boundaries than those without the MELF. In contrast, when the deep model is without the MELF, the features are blurred, especially in images with limited textures. The deep model struggles to effectively extract enough discriminative information for matching and registration, ultimately impairing registration performance. These visualization results show that our proposed MELF can encourage the deep module to explore rich features for more accurate registration. MELF employs various experts to extract image features from different perspectives and uses a learnable soft router for adaptive feature aggregation. Therefore, MELF can extract rich discriminative information from images with poor textures and significantly improve the registration performance.

\subsubsection{Visualization of registration results}
This part visualizes cross-modal image registration results and corresponding similarity maps of our proposal and F3Net, as shown in Fig. \ref{fig:visulization_registration_results}. In the optical image, the red and green boxes represent the ground-truth and predicted matching positions, respectively. We also show the similarity map of features across images, where the red and black dots indicate the ground-truth and predicted matching coordinates, respectively. 

Firstly, from the similarity maps, we can observe that F3Net tends to generate multiple peaks of high similarity responses, whereas our method suppresses similarity values in non-matching areas and maintains an unimodal similarity peak in the registration regions. Secondly, our method can accurately register images with weak and local repeated textures (see the fourth and fifth samples). F3Net produces multiple similarity peaks and finds a rough matching position. In contrast, our method predicts positions with smaller errors, even precisely locating the true position. This is because our method integrates global context information and local features, enhancing the feature discrimination and improving the registration performance of images with weak and local repeated textures. However, our proposal and F3Net should improve the registration performance in extreme and challenging cases. For example, in the sixth sample, we observe that the optical image contains few textures, and its local features are highly similar and repeatable, which will lead to multiple high responses in the similarity map and mismatching results. 

\subsection{Robustness Analysis}
In this section, we analyze the sensitivity of the proposed method to noise based on the OS dataset. The optical images in the OS dataset are added to different levels of Gaussian white noise with a mean of 0 and variance from 2\% to 30\%. We present their registration results of CMR (with $T=1$ and $T=3$) and average $L_{2}$. As shown in Fig. \ref{fig:robustness}, when the image noise variance increases from 2\% and 20\%, although our method is not trained on the noised images, the registration performance is relatively stable. When the Gaussian noise variance exceeds 22\%, the registration performance decreases slightly. These experimental results indicate that the proposed method is robust to noise. The main reason is that the Mamba-based model and the MELF enhance salient feature learning and handle challenging scenarios effectively. 

\section{Conclusion}
This paper proposes a novel multi-expert learning framework with the State Space Model (ME-SSM) for cross-modal remote sensing image registration. Firstly, ME-SSM proposes a multi-expert learning framework (MELF) to enhance feature extraction and improve image registration performance on images with limited textures. MELF leverages multiple feature experts to extract discriminative information from various transformed images and employs a learnable soft router to dynamically fuse the features from multiple experts. Secondly, we introduce the Mamba framework into image registration for the first time, which can efficiently capture global contextual features in images with slight computational complexity by employing a multi-directional cross-scanning strategy. Additionally, ME-SSM integrates an MFA module to further enhance multi-scale feature learning and facilitate effective interaction between global and local features, thereby improving registration accuracy. ME-SSM shows significant advantages in optical and SAR image registration, particularly for images with limited textures. However, the registration performance needs further improvement on the challenging images with very poor and highly repetitive textures. Additionally, remote sensing foundation models \cite{SelectionMAE, SARATR-X} have general feature representation capabilities, which can enhance shared feature learning from multi-modal images. In the future, we will explore integrating the foundation models with our MELF to improve cross-modal image registration performance.  

\bibliographystyle{IEEEtran}
\bibliography{references}

\begin{IEEEbiography}[{\includegraphics[width=1in,height=1.25in,clip,keepaspectratio]{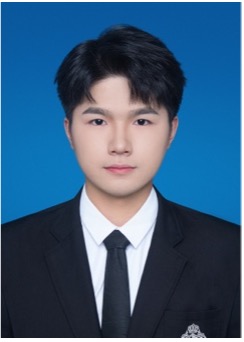}}]{Wei Wang} received the B.E. degree from Guilin University of Electronic Technology, Guilin, China, in 2023, and is currently pursuing a master's degree at the Key Laboratory of Intelligent Perception and Image Understanding, Ministry of Education of China. 
His research interests include deep learning, image matching, and cross-view geo-localization.
\end{IEEEbiography}

\begin{IEEEbiography}[{\includegraphics[width=1in,height=1.25in,clip,keepaspectratio]{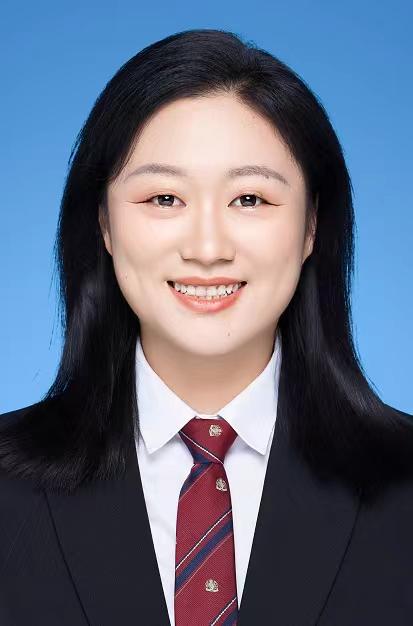}}]{Dou Quan} received the B.S. degree in 2015 and the Ph.D. degree in 2021, from Xidian University, Xi’an, China. She is currently an associate professor with the Key Laboratory of Intelligent Perception and Image Understanding of Ministry of Education of China, School of Artificial Intelligence, Xidian University.

Her research interests include machine learning, deep learning and metric learning, image matching, image registration, and image classification.
\end{IEEEbiography}

\begin{IEEEbiography} [{\includegraphics[width=1in,height=1.25in,keepaspectratio]{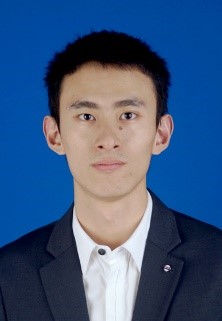}}]{Ning Huyan} received his B.S. degree in 2015 and Ph.D. degree in 2022 from Xidian University, Xi’an, China. From 2019 to 2020, he was a joint Ph.D. candidate under the supervision of Prof. Jocelyn Chanussot at the Inria Grenoble Rhône-Alpes research center in France. From 2022 to 2024, he worked as a researcher at Sensetime in Xi’an, China. Currently, he is a postdoctoral researcher in the Department of Automation at Tsinghua University. 

His research interests include hyperspectral image anomaly detection, human pose estimation, and motion generation.
\end{IEEEbiography}

\begin{IEEEbiography}[{\includegraphics[width=1in,height=1.25in,clip,keepaspectratio]{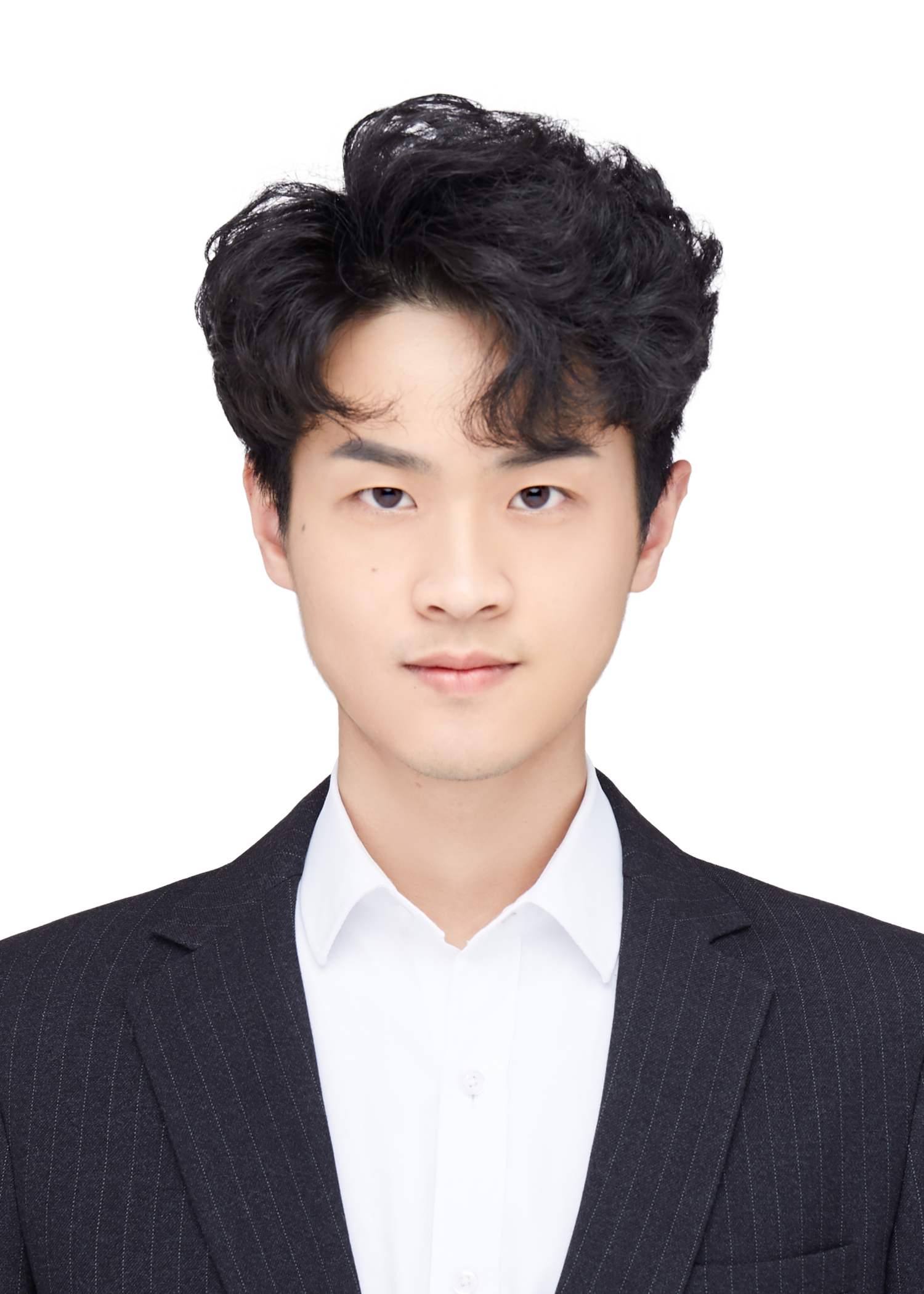}}]{Chonghua Lv} received the B.S. degree from Xi'an University of Posts and Telecommunications, Xi'an, China, in 2021, where he is pursuing the PhD. degree with the Key Laboratory of Intelligent Perception and Image Understanding, Ministry of Education of China, Xidian University. His research interests include deep learning, knowledge distillation, and semantic segmentation.
\end{IEEEbiography}

\begin{IEEEbiography} [{\includegraphics[width=1in,height=1.25in,keepaspectratio]{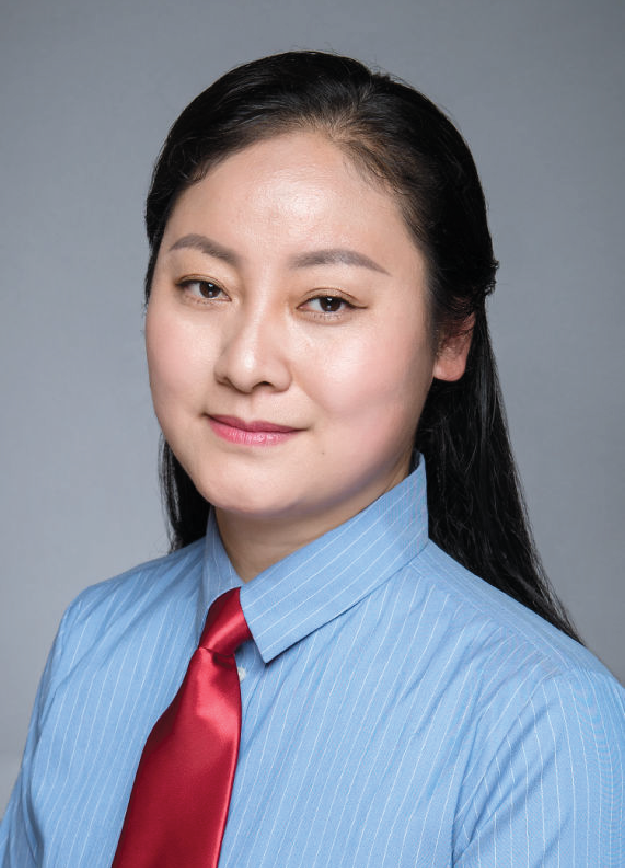}}]{Shuang Wang} received the B.S. degree in 2000, the M.S. degree in 2003, and the Ph.D. degree in circuits and systems in 2007 from Xidian University, Xi'an, China. She is currently a Professor with the Key Laboratory of Intelligent Perception and Image Understanding of Ministry of Education of China, Xidian University.

Her research interests include sparse representation, image processing, synthetic aperture radar (SAR) automatic target recognition, remote sensing image captioning, and polarimetric SAR data analysis and interpretation.
\end{IEEEbiography}

\begin{IEEEbiography}[{\includegraphics[width=1in,height=1.25in,clip,keepaspectratio]{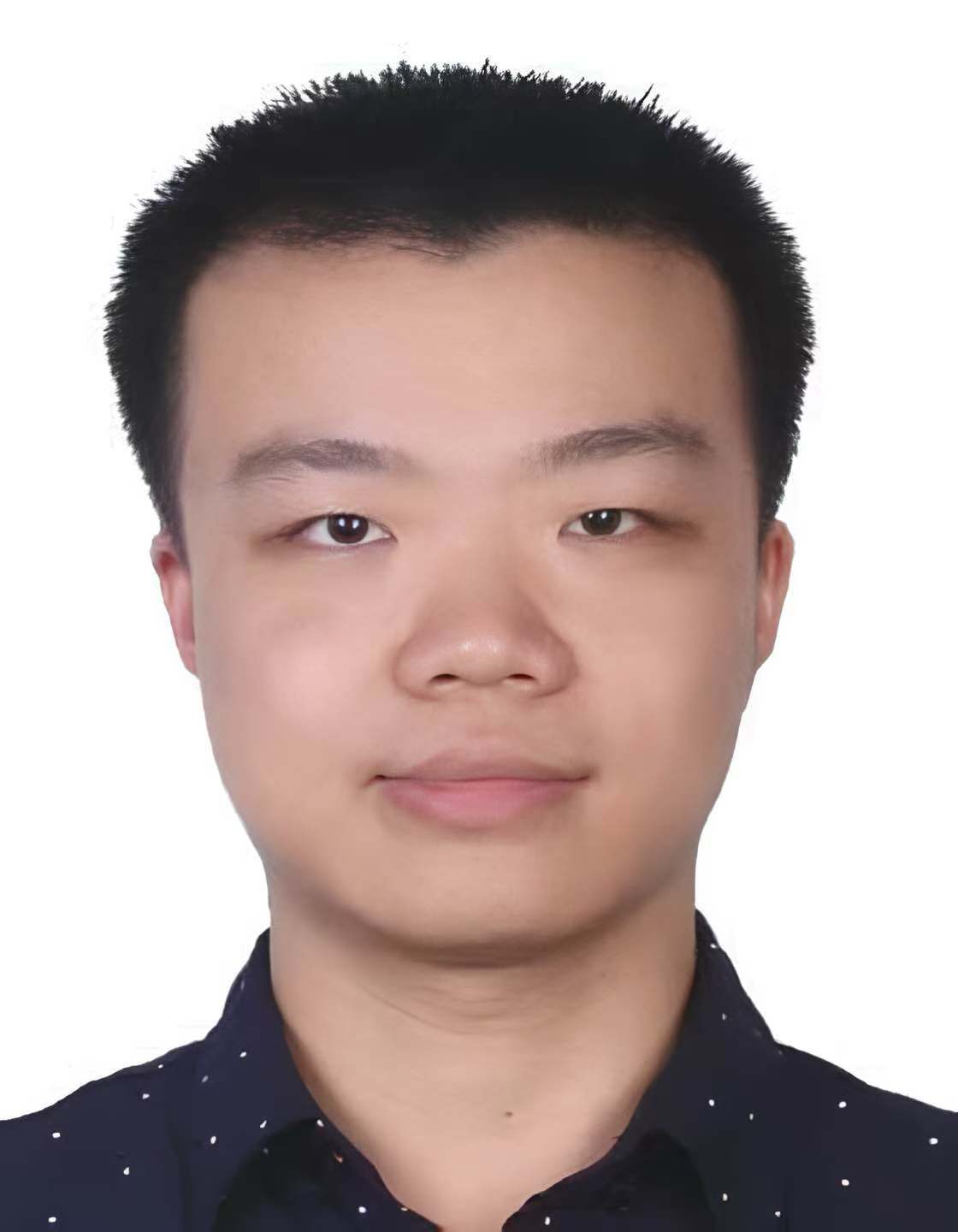}}]{Yunan Li} received his B.S. and Ph.D. degrees from the School of Computer Science and Technology, Xidian University, Xi'an, China in 2014 and 2019, respectively. 

He is an Associate Professor at Xidian University. His research interests include computer vision and pattern recognition, especially their applications in image enhancement and action/gesture recognition.
\end{IEEEbiography}

\begin{IEEEbiography}[{\includegraphics[width=1in,height=1.25in,clip,keepaspectratio]{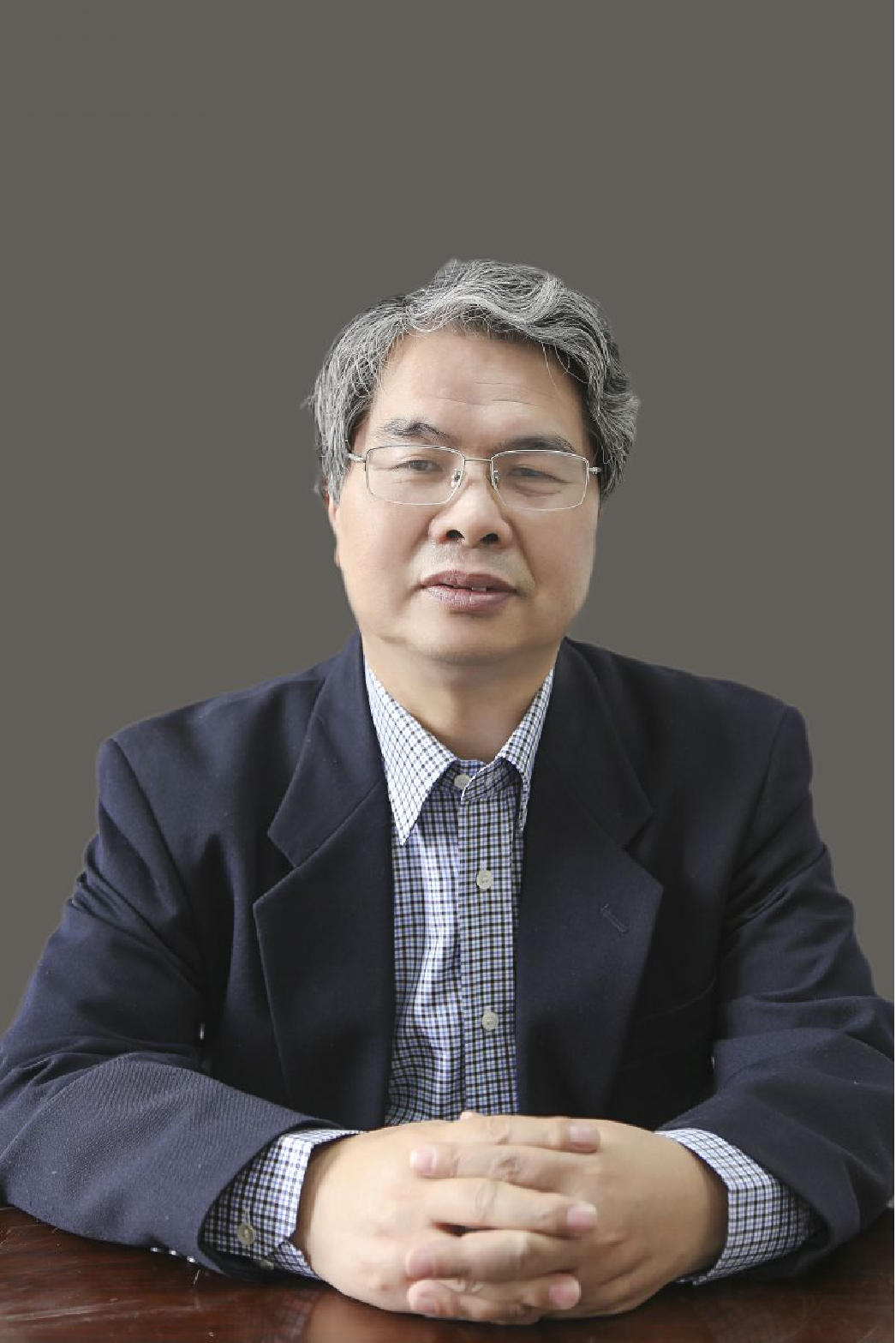}}]{Licheng Jiao}
	was born in Shaanxi, China, on October 15, 1959. He received the B.S. degree from Shanghai Jiaotong University, China, in 1982 and the M.S. and Ph.D. degrees from Xi'an Jiaotong University, Xi'an, China, in 1984 and 1990, respectively. From 1984 to 1986, he was an Assistant Professor with the Civil Aviation Institute of China, Tianjin, China. During 1990 and 1991, he was a Postdoctoral Fellow with the Key Lab for Radar Signal Processing, Xidian University, Xi'an, China. Currently, he is the Director of the Key Laboratory of Intelligent Perception and Image Understanding of Ministry of Education of China.
	
	His current research interests include signal and image processing, nonlinear circuits and systems theory, learning theory and algorithms, optimization problems, wavelet theory, and machine learning.
\end{IEEEbiography}
\end{document}